\documentclass[acmtog]{acmart}

\makeatletter
\let\@authorsaddresses\@empty
\makeatother

\renewcommand\footnotetextcopyrightpermission[1]{} 

\usepackage{booktabs} 

\usepackage{caption}
\usepackage{subcaption}
\usepackage{appendix}

\usepackage{amsmath,bm}

\newcommand{\Edet}{E_\mathrm{Det}}
\newcommand{\Erecon}{E_\mathrm{Recon}}
\newcommand{\Edist}{E_\mathrm{Dist}}
\newcommand{\Esmooth}{E_\mathrm{Smooth}}
\newcommand{\Edsmooth}{E_\mathrm{DSmooth}}
\newcommand{\Epull}{E_\mathrm{Pull}}

\usepackage[ruled]{algorithm2e} 

\SetAlFnt{\small}
\SetAlCapFnt{\small}
\SetAlCapNameFnt{\small}
\SetAlCapHSkip{0pt}

\DeclareMathOperator{\argmin}{argmin} 

\SetCommentSty{mycommfont}
\SetKw{Continue}{continue}
\SetKw{Break}{break}

\setcopyright{none}
\begin{document}
\title{Garment Avatars: Realistic Cloth Driving using Pattern Registration}

\author{Oshri Halimi}
 \affiliation{%
 \institution{Technion – Israel Institute of Technology}
 \country{Israel}}
\affiliation{%
 \institution{Facebook Reality Labs}
 \country{USA}}
\email{oshri.halimi@gmail.com}

\author{Fabian Prada}
\affiliation{%
 \institution{Facebook Reality Labs}
 \country{USA}}
\email{fabianprada@fb.com}

\author{Tuur Stuyck}
\affiliation{%
 \institution{Facebook Reality Labs}
 \country{USA}}
\email{tuur@fb.com}

\author{Donglai Xiang}
\affiliation{%
 \institution{Carnegie Mellon University}
 \country{USA}}
\affiliation{%
 \institution{Facebook Reality Labs}
 \country{USA}}
\email{donglaix@cs.cmu.edu}

\author{Timur Bagautdinov}
\affiliation{%
 \institution{Facebook Reality Labs}
 \country{USA}}
\email{timurb@fb.com}

\author{He Wen}
\affiliation{%
 \institution{Facebook Reality Labs}
 \country{USA}}
\email{hewen@fb.com}

\author{Ron Kimmel}
\affiliation{%
 \institution{Technion – Israel Institute of Technology}
 \country{Israel}}
\email{ron@cs.technion.ac.il}

\author{Takaaki Shiratori}
\affiliation{%
 \institution{Facebook Reality Labs}
 \country{USA}}
\email{tshiratori@fb.com}

\author{Chenglei Wu}
\affiliation{%
 \institution{Facebook Reality Labs}
 \country{USA}}
\email{chenglei@fb.com}

\author{Yaser Sheikh}
\affiliation{%
 \institution{Facebook Reality Labs}
 \country{USA}}
\email{yasers@fb.com}

\begin{abstract}
Virtual telepresence is the future of online communication. 
Clothing is an essential part of a person's identity and self-expression. 
Yet, ground truth data of registered clothes is currently unavailable in the required resolution and accuracy for training telepresence models for realistic cloth animation. 
Here, we propose an end-to-end pipeline for building drivable representations for clothing.
The core of our approach is a multi-view patterned cloth tracking 
algorithm capable of capturing deformations with high accuracy.
We further rely on the high-quality data produced by our tracking method to build a \textit{garment avatar}: an expressive and fully-drivable geometry model for a piece of clothing.
The resulting model can be animated using a sparse set of views and produces highly realistic reconstructions which are faithful
to the driving signals. 
We demonstrate the efficacy of our pipeline on a realistic
virtual telepresence application, where a garment is being reconstructed from two views, and a user can pick and swap garment
design as they wish.
In addition, we show a challenging scenario when driven
exclusively with body pose, our drivable garment avatar is capable of producing realistic cloth geometry of significantly
higher quality than the state-of-the-art.  
\end{abstract}

\begin{teaserfigure}
\centering 
\includegraphics[width=\textwidth]{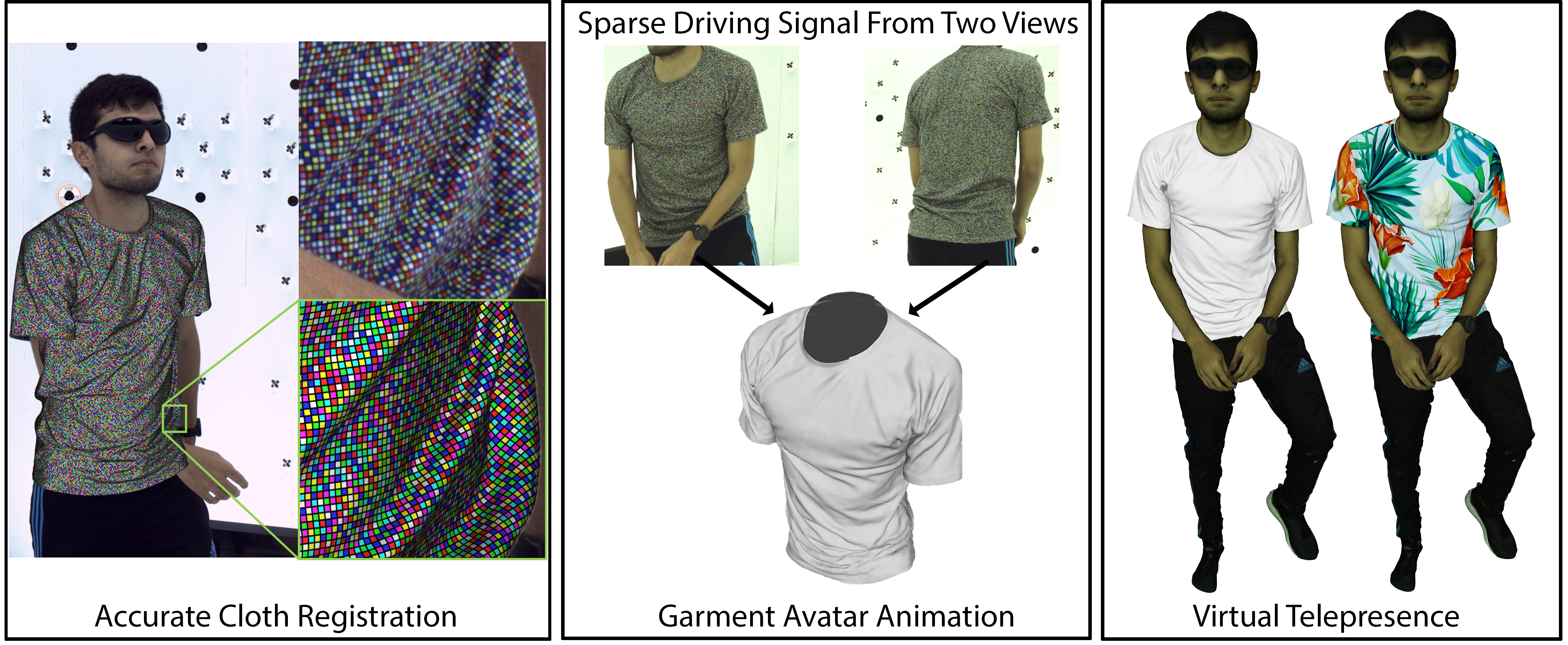} 
\caption{\textbf{Left:} Comparison of our cloth registration and capture image. \textbf{Middle:} Garment Avatar animated from two cameras.
\textbf{Right:} Virtual telepresence with our drivable garment avatar with texture editing.} 
\label{fig:teaser} 
\end{teaserfigure}

\maketitle

%
%


%
%

\keywords{Garment Capture, Telepresence, Virtual Clothing, Machine Learning, Computer Vision, Registration}

\section{Introduction}

Truthful representation of humans is essential for virtual telepresence applications, the future of communication and avatars in the metaverse. 
Previous efforts have focused primarily on human heads~\cite{lombardi2018deep} and bodies~\cite{bagautdinov2021driving,peng2021neural}. 
Due to the highly complicated motion of garments, it still remains elusive to teleport the clothing faithfully on a body avatar. 
People consciously pick out their outfits to follow fashion trends and 
accurately conveying the clothing is a part of ones identity and should be accounted for to obtain satisfactory telepresence experiences. 

One major challenge in representing clothing comes from the lack of 
high-quality cloth registration of the stretching and shearing of the fabric on moving bodies, which is notoriously difficult due to the numerous 
self-occlusions and the lack of visual features for alignment. 
Previous work have made efforts in capturing simple cloth swatches under external forces in a controlled environment~\cite{miguel2012data, rasheed2020learning, clyde2017modeling, wang2011data, bhat2003estimating}. 
However, these captures only consider isolated suspended fabrics without capturing the combined effects of friction, air drag, external forces or the interaction of garments being worn with the underlying body.
Therefore, ground truth data of registered clothes is still unavailable 
in the required resolution and accuracy, limiting the quality of drivable garment 
models~\cite{xiang2021modeling}. 
To this end, we propose a novel cloth registration pipeline that can operate in a multi-view camera setting and produces a dynamic sequence of registered cloth meshes with an unprecedented registration resolution of 2.67 millimeters and triangulation localization error of 1 millimeter.
This registration method relies on a novel pattern printed on the cloth, designed to optimize the ratio of pixel area to the number of uniquely registered cloth surface points. 
We localize dense pattern primitives using a specialized image-based 
detector and identify each pattern keypoint with a graph processing algorithm that robustly handles the combination of non-rigid and projective transformations, self-occlusions, and detection noise. 
%

Another limitation of cloth teleporting emerges from the driving mechanism itself; While some drivable garment models \cite{xiang2021modeling,habermann2021real} use the body pose as the driving signal, the pose doesn't map uniquely to the garment geometry and appearance. 
As a result, pose-dependent models tend to either 
significantly smooth out details due to insufficient conditioning, or introduce high-frequency deformations
which are not faithful to the true underlying clothing state due to over conditioning. Although our high-accurate registered data 
would already benefit these methods, we additionally propose a novel animatable 
clothing model - a garment avatar - that fully takes advantage of our 
high-quality registered data for a much higher fidelity driving result of the garment.

%
%
%
More specifically, we propose utilizing the partial pattern detection as the driving signal, acquired
by applying our registration method on two extremely sparse views. Although only partially available, this driving signal has much more spatial correlations with the whole garment state, thus achieving better fidelity.  
%
As two driving views are selected to be non-overlapping to achieve the best coverage, the traditional triangulation would not work here. To solve this, we formulate the whole reconstruction problem as an coordinate inpainting problem by designing an inpainting UNet-type network
that takes as input multiple partial pixel coordinate signals
in UV space, and produces complete signal in 3D.
%
Our key insight comes from the fact that if one can inpaint all incomplete
2D observations each view, then one can lift those to 3D with triangulation.
%
To allow for better generalization, both inputs and outputs
of the model are defined with respect to a coarse 
kinematic model fitted to the partial observations.
Notice that, our kinematic model is agnostic to the body shape, 
and ultimately can be applied to arbitrary type of garments.

In our experiments, we demonstrate that our garment tracking pipeline
performs favorably to the state-of-the-art both qualitatively and
quantitatively.
In particular, our results illustrate that the data produced by our 
novel cloth tracking method significantly boosts performance of
state-of-the-art animatable 
garment models driven from pose.
Finally, we show the efficacy of our drivable garment avatar model
on a realistic virtual telepresence application, where the garment
worn by a user is fully reconstructed from two RGB views, 
and a user can pick and swap garment design on-the-fly.

In summary, our main contributions are:
\begin{itemize}
    \item We develop a carefully designed cloth registration pipeline that captures cloth in high accuracy and dense correspondences.
    \item We develop a method for accurate and realistic cloth animation from pixel registration obtained from two camera video streams.
    \item We demonstrate that the training data obtained from our registration method can significantly improve the output quality when applied to pose-driven animation.
\end{itemize}

\section{Related Work}
\subsection{Multi-View Clothing Capture}

Multi-view clothing capture has been explored as a source of geometry for garment modeling. A typical multi-view clothing capture pipeline consists of three steps, geometry reconstruction, clothing region segmentation, and registration.

In the geometry reconstruction step, typically the raw surface of the clothed human body is reconstructed from RGB input images by multi-view images using Multi-View Stereo (MVS) or from 3D scanners using Photometric Stereo. In the segmentation step, the garment region is identified and segmented out. Early work \cite{bradley2008markerless,Pons-Moll:Siggraph2017} uses the difference of color between the garment and the skin as the primary source of information for segmentation, while more recent work \cite{bang2021estimating,xiang2021modeling,bhatnagar2019multi} aggregates clothing parsing results from multi-view RGB images to perform the segmentation more robustly.

Compared with the reconstruction and segmentation steps, the registration of garment is a more open question and also the focus of this work. The goal of registration is to represent the complete geometry of the clothing in different frames with a fixed mesh topology and encode the correspondences by vertices. Previous literature of clothing registration falls into the following two categories: \textbf{template-based} methods and \textbf{pattern-based} methods.

\textbf{Template-based registration} methods start from a pre-defined template topology and fit the template to each instance of the captured garment according to the geometry. Early work \cite{bradley2008markerless} attempts to find consistent cross-parameterization among different frames of the garment from a common base mesh by minimizing the stretching distortion, and then explicitly representing the correspondences by re-triangulation.

The majority of work in this category \cite{zhang2017detailed,ma2020cape,xiang2020monoclothcap,Pons-Moll:Siggraph2017,bhatnagar2019multi,tiwari2020sizer,xiang2021modeling} uses non-rigid Iterative Closest Point (ICP) to fit template mesh to the target clothing geometry. The problem is formulated as a minimization of surface distance between the free-form template and the target, plus some regularization term that preserves the quality of mesh triangulation. To provide better initialization for the optimization, some work \cite{zhang2017detailed,ma2020cape,xiang2020monoclothcap} first uses a kinematic model such as SMPL \cite{loper2015smpl} to help estimate a coarse human surface, which is further aligned with the reconstructed clothing shape by allowing free-form deformation in the SMPL body topology. Some recent work \cite{bhatnagar2020combining,bhatnagar2020loopreg} replaces the explicit optimization with the prediction from a neural network to avoid the difficulty of robustly initializing and regularizing the optimization. The above SMPL+D formulation assumes a one-to-one fixed correspondence between the body template and the clothing, which is often violated due to tangential relative movement between the fabric and the body, as well as the invisible clothing regions in the wrinkle folds.

Therefore, some work \cite{Pons-Moll:Siggraph2017,bhatnagar2019multi,tiwari2020sizer,xiang2021modeling} further separate the presentation of the clothing from the underlying body, and use the segmented boundary to guide the deformation. While these methods can produce visually appealing registered clothing sequences, they suffer from the fundamental limitation of inferring correspondences purely from geometry. There are no explicit clues for the correspondences between frames except the regularization of mesh triangulation or vertex distances. Therefore, the registration output generally suffer from correspondence errors since there is no mechanism to ensure that each vertex coherently tracks the same physical point.


By comparison, \textbf{pattern-based registration} methods, which our approach belongs to, do not suffer from the ambiguity in correspondences due to lack of visual cues as in template-based registration methods. The key idea is to use identifiable pattern to explicitly encode correspondences on the captured surface. Similar concepts have been widely explored in the use of checker boards for camera calibration \cite{dao2010robust}, and the application of fiducial markers, like ARTag \cite{fiala2005artag}, AprilTag \cite{olson2011apriltag, wang2016apriltag} and ArUco \cite{garrido2014automatic}.

Specific to the area of garment capture, early work \cite{pritchard2003cloth,scholz2005garment,white2007capturing} utilizes classical computer vision techniques such as corner detection and multi-view geometry to reconstruct and identify printed markers on the garments. With the help of the pattern, the correspondences on the garments can be robustly tracked in visible sections and reliably estimated in regions occluded by folds and wrinkles. Our work extends the color-coded pattern approach \cite{scholz2005garment} to achieve denser detection.

In recent years, the constantly evolving frontier of learning-based computer vision algorithms enables revisiting this research problem with a plethora of enhanced image processing capabilities. The focus has shifted to the pattern design question and its resulting information theory properties. Specifically, to allow high-resolution capture, one should design the pattern to detect as many as possible points per surface area. For example, Yaldiz and colleagues \cite{CNNMarker:SIG:2021} generate learnable fiducial markers optimized for robust detection under surface deformations, using a differentiable renderer for end-to-end training. To make the marker detection resilient to surface deformations, they use geometric augmentations such as radial and perspective distortions and TPS (thin-plate-spline) deformation. Those kinds of deformation are applied in the 2D image space and cannot simulate self-occlusions resulting from the 3D folding of the surface. Currently, fiducial-markers-based tracking methods cope with relatively mild deformations and still do not address complex deformations containing folds and wrinkles. 

In the clothing registration problem, the pursuit of high resolution makes characters and symbols unsuitable for the printed pattern due to their low density of correspondences. Thus Chen and colleagues \cite{chen2021capturing} propose to identify a corner from its immediate surrounding squares printed on a tight suit worn by the subjects. In our setting each center might cover no more than a few pixels so we must rely on simpler attributes that can be robustly detected from the low resolution observation of the board squares in the images.

\subsection{Clothing Animation}

Our end goal in this paper is to build a clothing model that can be animated, or driven, from very sparse input driving signals (binocular RGB video streams in our case) for the purpose of VR telepresence. Therefore, an ideal system should satisfy two requirements: first, the animation output should faithfully reflect the current status of the clothing captured from the input driving signal; second, it should fully utilize prior knowledge about the appearance of the clothing that is being teleported.

\textbf{Pose-driven clothing animation} aims to produce realistic clothing animation from the input pose represented by 3D joint angles, or the underlying body skinned according to the joint angles by a kinematic model. \emph{Physics-based simulation} of garments \cite{baraff1998large,narain2012adaptive,stuyck2018cloth} has been studied for a long time and established as a standard approach for clothing animation in the movie and gaming industry. In recent years, there has been a lot of interest in using \emph{data-driven} approaches, especially deep neural networks, to directly learn clothing animation from data paired with the input body poses \cite{lahner2018deepwrinkles,ma2020cape,santesteban2021self,saito2021scanimate,SCALE:CVPR:2021,POP:ICCV:2021,xiang2021modeling,10.1145/3478513.3480479,habermann2021real}. While these approaches can produce visually appealing animation, the input body motion alone does not contain enough information to guarantee the consistency between the animation output and the real clothing status of the teleported subject, as these pose-driven approaches do not utilize any visual cues of the current appearance.

Our binocular clothing animation setting is also related to the recent work in \textbf{performance capture} from monocular or sparse multi-view inputs, which can also serve the purpose of clothing animation for VR telepresence. This line of work can be further divided into two categories: \emph{shape regression} approaches and \emph{template deformation approaches}. The \emph{shape regression} approaches train deep neural networks to regress per-frame clothed human shape from monocular \cite{natsume2019siclope,saito2019pifu,saito2020pifuhd,alldieck2019tex2shape,li2020monocular} or sparse multi-view inputs \cite{huang2018deep,bhatnagar2019multi}. These approaches enjoy the flexibility of being able to address different subjects and clothing with a single network, but are usually limited in output quality due to the fundamental ambiguity of inferring complicated clothing geometry without prior knowledge about the specific subject. More related to our work are the \emph{template deformation} approaches. These approaches utilize a pre-scanned personalized template of a specific subject to track the clothing deformation from a single RGB video \cite{xu2018monoperfcap,habermann2019livecap,habermann2020deepcap,habermann2021deeper,li2021deep}. The templates may also be built on-the-fly by fusing different frames of geometry from RGB-D input \cite{yu2018doublefusion,su2020robustfusion,yu2021function4d}. The personalized templates can provide strong prior knowledge to alleviate the 3D garment shape ambiguity. In this work, our patterned cloth serves as a special template whose correspondences can be easily inferred from the input driving signal, thus amenable to high-quality clothing animation in VR telepresence.

\section{Outline}
\begin{figure}
     \centering
     \begin{subfigure}[b]{0.5\textwidth}
         \centering
         \includegraphics[width=\textwidth]{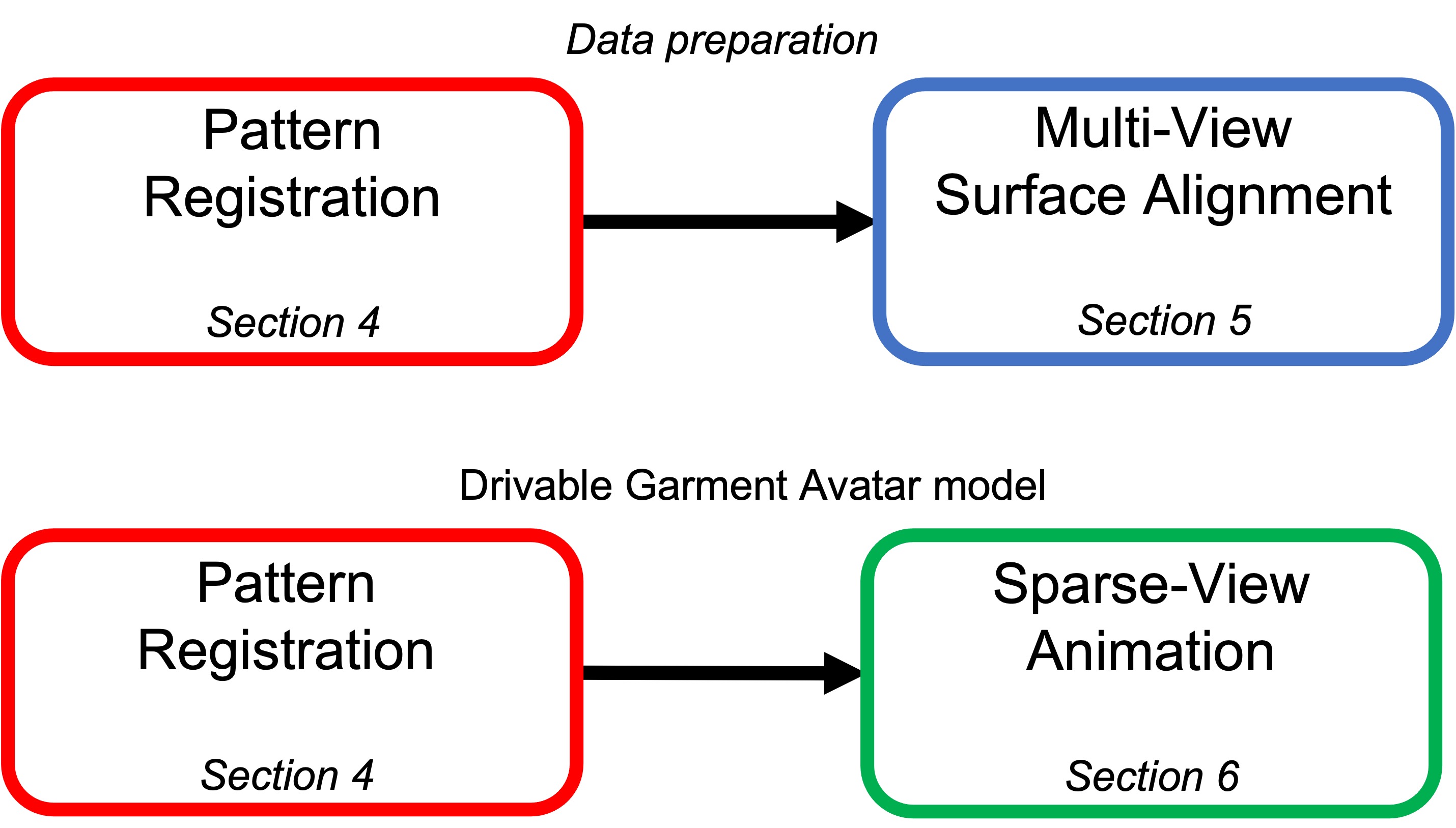}
         \caption{\textbf{Method Overview.} The data preparation process (top) reconstructs the registered cloth surface in a multi-view setting.
We use the high-quality cloth registration data to train our drivable Garment Avatar model that animates the cloth, utilizing a limited number of cameras}
         \label{fig:diagram}
     \end{subfigure} \vfill \vfill \vfill
     \begin{subfigure}[b]{0.5\textwidth}
         \centering
         \includegraphics[width=\textwidth]{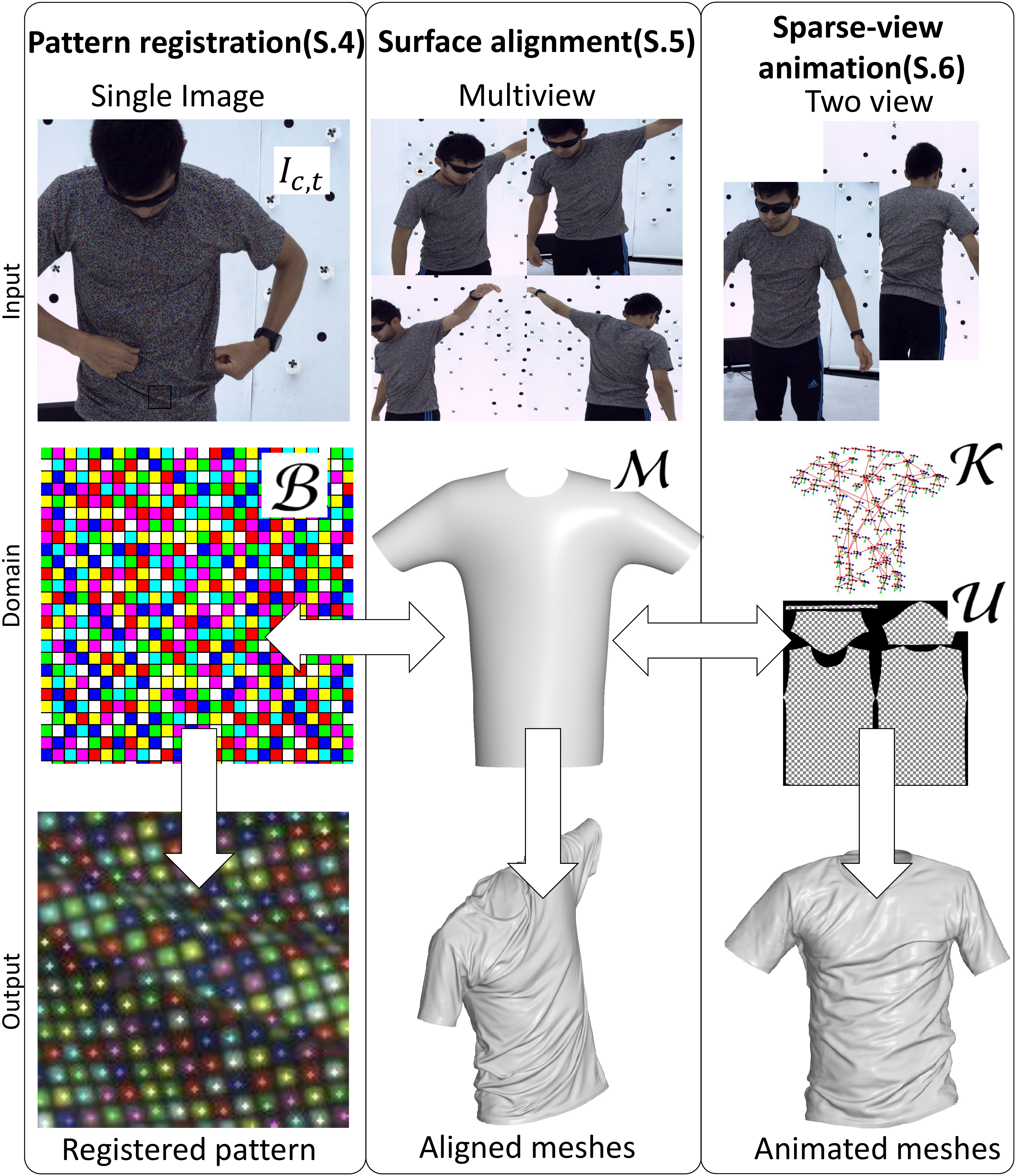}
         \caption{Illustration of the three modules in the above diagram. For a given camera $c$, and time step $t$: the \textbf{Pattern Registration} module registers the visible pattern cells in a single image $I_{c, t}$; Each registration has two attributes 1) pixel coordinates in the image space 2) grid coordinates in the pattern board domain $B$. The \textbf{Multi-View Surface Alignment} stage aligns a template mesh $M$ according to the multi-view image registrations. The \textbf{Sparse-View Animation} module animate the cloth from the image registrations of two cameras by mapping the registrations to the UV domain, $U$, and inpainting the partial signals. A crucial step in the animation fits a coarse surface to the driving camera registrations, represented by a surface kinematic model $K$.}
         \label{fig:overview}
     \end{subfigure}
        \caption{Overview of the three main components in our drivable Garment Avatar model}
        \label{fig:methodOverview}
\end{figure}

This work aims to build an expressive and drivable neural garment model, 
capable of generating faithful and realistic cloth animations. 
To this end, we created our high-quality cloth registration data to train our novel cloth driving model. 
The global overview of our method is provided in Figure~\ref{fig:diagram} depicting the data preparation and cloth driving stages. Figure \ref{fig:overview} includes an overview for each of the three modules that appear in the general diagram.
%

\begin{figure*}
\centering 
\includegraphics[width=0.9\textwidth]{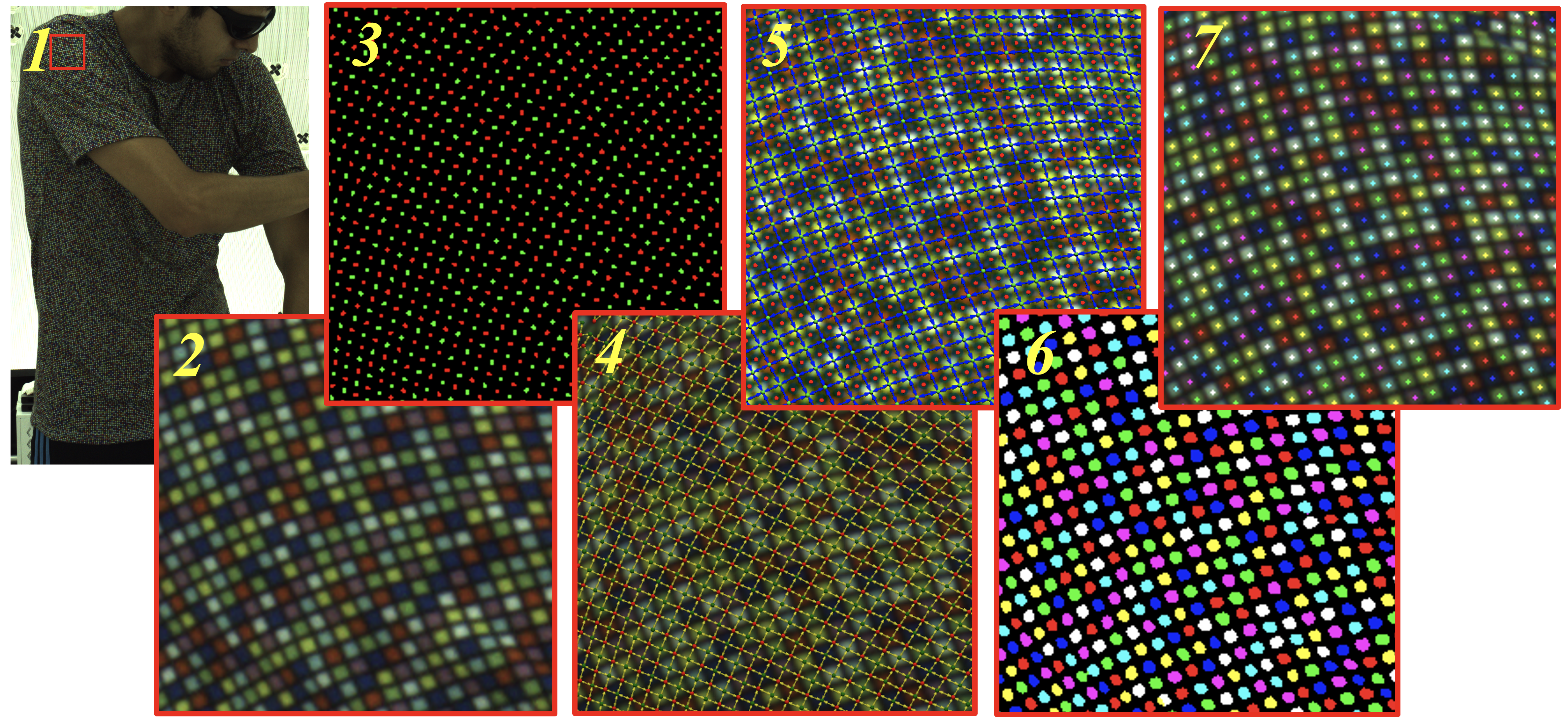}
\caption{\textbf{Patterned Cloth Registration.} 1) Original frame 2) Zooming in on the pattern 3) keypoint detection by \textit{SquareLatticeNet} - corners (red) and centers (green) 4) heterogeneous graph 5) homogeneous graph 6) color detection by \textit{ColorBitNet} - pixel classification 7) Registered pixels: every pixel detection is assigned with coordinates in the pattern domain $B$.} 
\label{fig:pattern-registration}
\end{figure*}

The first stage is \textbf{Pattern Registration}, described in Section~\ref{sec:pattern-registration}, see 
Figure~\ref{fig:pattern-registration}, is the shared component between both model building and model animation.
Given a single frame capturing a performer wearing a grid-like patterned garment, we register every visible pattern cell, with predefined grid coordinates in the pattern domain $B$, to pixel coordinates in the image space.

In the \textit{data preparation} branch, we use the registered frames obtained from a multi-view camera system as an input to our \textbf{Multi-View Surface Alignment} stage, which we describe in Section~\ref{sec:panoptic_registration} and visualize in Figure~\ref{fig:panoptic_registration}. First, we triangulate all the registered frames to get registered point clouds where every point maps to specific coordinates in the pattern domain $B$. Then, we align a mesh template $M$ to the resulting point clouds. 

In the \textit{Garment Avatar} model branch, we use a sparse set of registered frames as the driving signal to our \textbf{Sparse-View Animation} module, described in Section~\ref{sec:animation}, see Figure~\ref{fig:animation}. The animation module consists of our \textit{Pixel Driving} network and a coarse geometry estimation procedure. Specifically, the \textit{Pixel Driving} receives the pixel location signal detected per camera, as defined in the UV space, and outputs the world space 3D coordinates signal in the same UV domain. 
Crucially, to ensure generalization to newly seen poses at inference time, we define the inputs and outputs to our network relative to a coarse geometry fitted to our sparse-view pixel registrations.

\section{Patterned Registration}
\label{sec:pattern-registration}

To enable accurate cloth captures, we manufacture a piece of cloth with a fine-grained color-coded pattern. We follow a similar approach to pattern desing as described by Scholz et al. \shortcite{scholz2005garment}. 
We introduce novel methods for robust registration of color-coded patterns that enables dense alignment of cloth.
\subsection{Pattern Design}

The pattern consist of a colored board where cells takes one of seven colors.
Cells are separated by grid lines to improve contrast at edges and corners of the board. We assign colors to cells in such a way that the color configuration on each $3\times3$ cell-set is unique, including w.r.t. board rotations. We further impose adjacent cells to have different colors to improve color disambiguation. 

\begin{figure}[h]
\centering 
\includegraphics[width=0.4\textwidth]{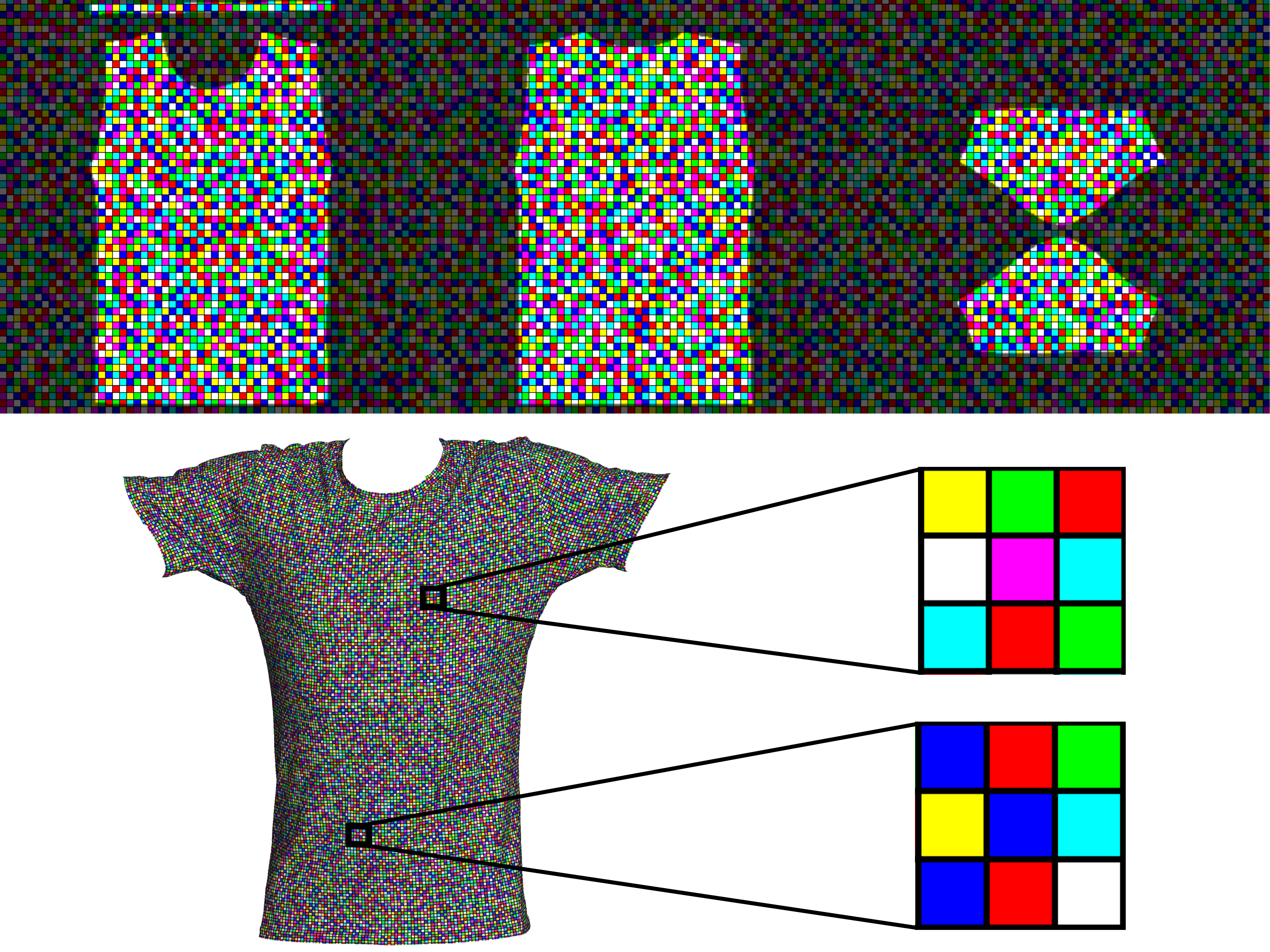}
\caption{\textbf{Pattern design and fabrication}. The colored board is printed on fabric (top) to manufacture a t-shirt (bottom left). Each $3\times3$ cell of the pattern (bottom right) has unique color configuration.} 
\label{fig:desing_and_fabrication} 
\end{figure}

We print a color board with $300\times 900$ cells on polyester fabric with a resolution of 2.7mm per cell. A t-shirt is manufactured by sewing the cut garment panels extracted from the fabric, see Figure~\ref{fig:desing_and_fabrication}.
Our manufactured t-shirt model contains 98618 cells.

Our choice of primitive colors and cell configuration balances locality, distinctiveness and uniqueness. A $3\times3$ cell-set provides us with a good match between locality and uniqueness: for a cell of 2.7mm length, we can uniquely localize a pattern element when the 8mm$\times$8mm patch around it is visible. Getting unique detections for cell-set smaller than $3\times3$ can be achieved by increasing the number of color primitives, however this reduce the distinctiveness of the color primitives, increasing quantization errors, making the pattern registration less robust.

\subsection{Image Detection}
\begin{figure}[h]
\centering 
\includegraphics[width=0.47\textwidth]{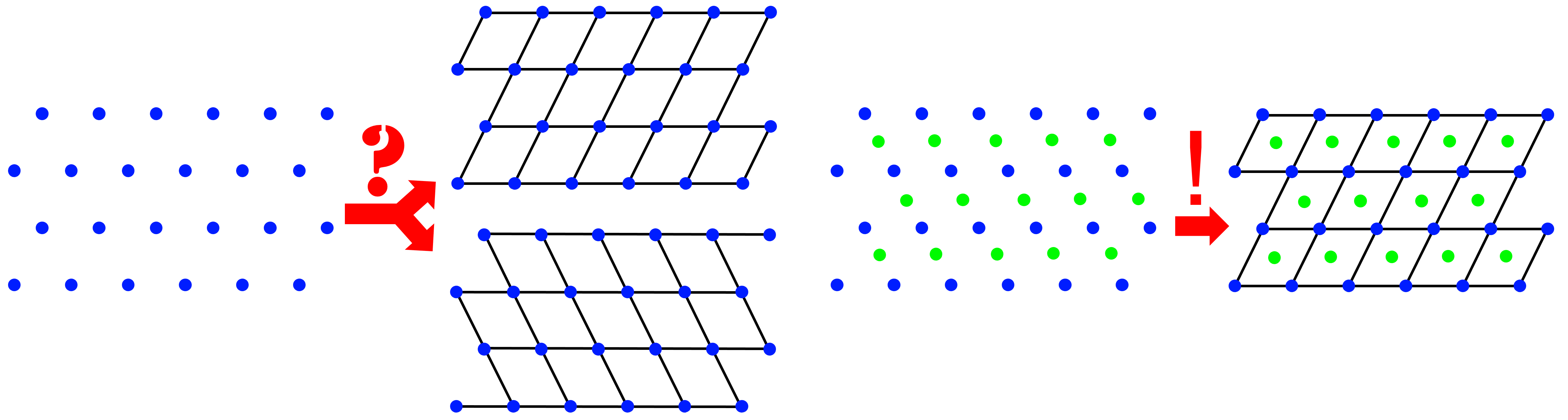}
\caption{Lattice neighborhood inference with squares and corners detection} 
\label{fig:lattice} 
\end{figure}

We propose an image pattern detector, \emph{PatterNet}, consisting of two separate networks. \emph{SquareLatticeNet} detects the corners and the centers. \emph{ColorBitNet} classifies the pixel color. Both networks have been implemented using a UNet.
%
%
The location of the square center can reveal valuable information about the topology; however, under a general affine transformation, it leads to a non-unique graph, as shown in Figure~\ref{fig:lattice}, due to the lattice symmetry.
To restrict the ambiguity, we additionally detect the square corners. 
We empirically observed that in our setting, the combination of centers
and corners is sufficient for our algorithm to determine the correct local neighborhood of each visible square in the image in most cases. 
Graph ambiguities are usually caused by potential high-skew transformations that create aliasing, and by adding the corners, we make the aliasing less frequent.
For a detailed review of the image pattern detector we refer the reader to the supplementary material.

\subsection{Graph Processing}
\label{sec:graph_processing}
This stage reconstructs the graph topology of the square grid from the center and corner detections obtained by the previous step. 
The detection contains a certain amount of noise, such as outliers, and parts of the graph are not visible in the image due to self-occlusion. 
Additionally, near the invisible graph parts, it's common to see discontinuities in the periodic structure of the detected lattice points, where nodes disconnected in the graph become proximate in the image. 
Our approach is to recover the heterogeneous graph defined on the square grid, utilizing the detected corners information, and construct the homogeneous grid graph from the former. Please see the supplementary material for a detailed description of both graph generation algorithms.
Our approach has the following advantages:
\begin{enumerate}
\item We have precise geometric requirements between opposite type neighbor, namely corners and centers, derived from the locally affine approximation that we can utilize to filter outlier detection and outlier graph edges.

\item Neighbors in the heterogeneous graph fall closer in the image than neighbors in the homogeneous graph. 
As a result, we can apply geometric regularizations between neighbors by neglecting the non-rigid deformation and assuming a locally affine transformation. 
    
\item Once the centers are registered, the heterogeneous graph enables the registration of the corners and the later triangulation, thus doubling the resolution of the registered cloth surface. We didn't exploit the last advantage since corner registration introduces additional steps into our registration pipeline and template generation, which we defer to future work.
\end{enumerate}

\subsection{Hash Code Inference}

This stage aims to calculate the hash code for every node in the homogeneous graph and to register the node to a board location using this code. 
After the previous stage, we have a $3\times 3$ grid-graph attribute around the center of each node in the homogeneous graph. Each node has a color attribute that is either a single color for confident detections or several candidate colors for ambiguous results. 
%
We are robust against possible color ambiguities and additionally offer a way to reject codes where the color is not ambiguous but the detection is wrong.
We propose the neighbor-voting-based hash extraction algorithm that enables information recovery given these incorrect detections.
Most code errors produce an empty result in the hash query.
However, correct codes always have a unique match on the board, and neighborhood relations are preserved from the homogeneous graph to the patches in the board domain, whereas this is not the case for incorrect codes. 
With this insight, we suggest the \textit{neighbor voting} algorithm to help recover ambiguous codes and to reject faulty but unambiguous codes. The algorithm is described in the supplementary and visualized in Figure~~\ref{fig:voting}.

\begin{figure}[h]
\centering 
\includegraphics[width=0.5\textwidth]{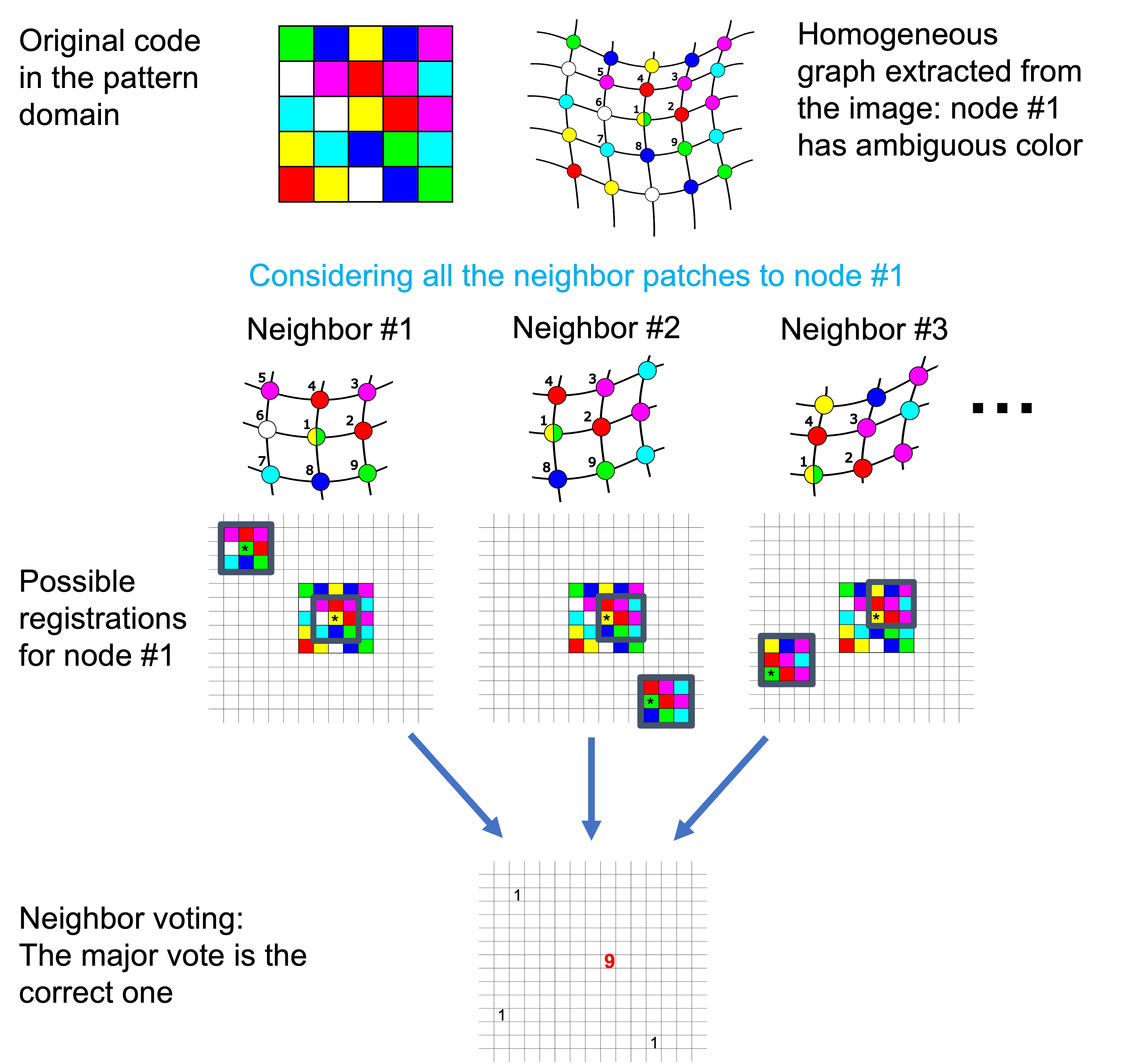}
\caption{\textbf{Neighbor voting algorithm illustration} A single color-bit ambiguity is resolved by the majority vote.} 
\label{fig:voting}
\end{figure}

\section{Multi-view Surface Alignment}
\label{sec:panoptic_registration}
\begin{figure*}
\centering 
\includegraphics[width=1.0\textwidth]{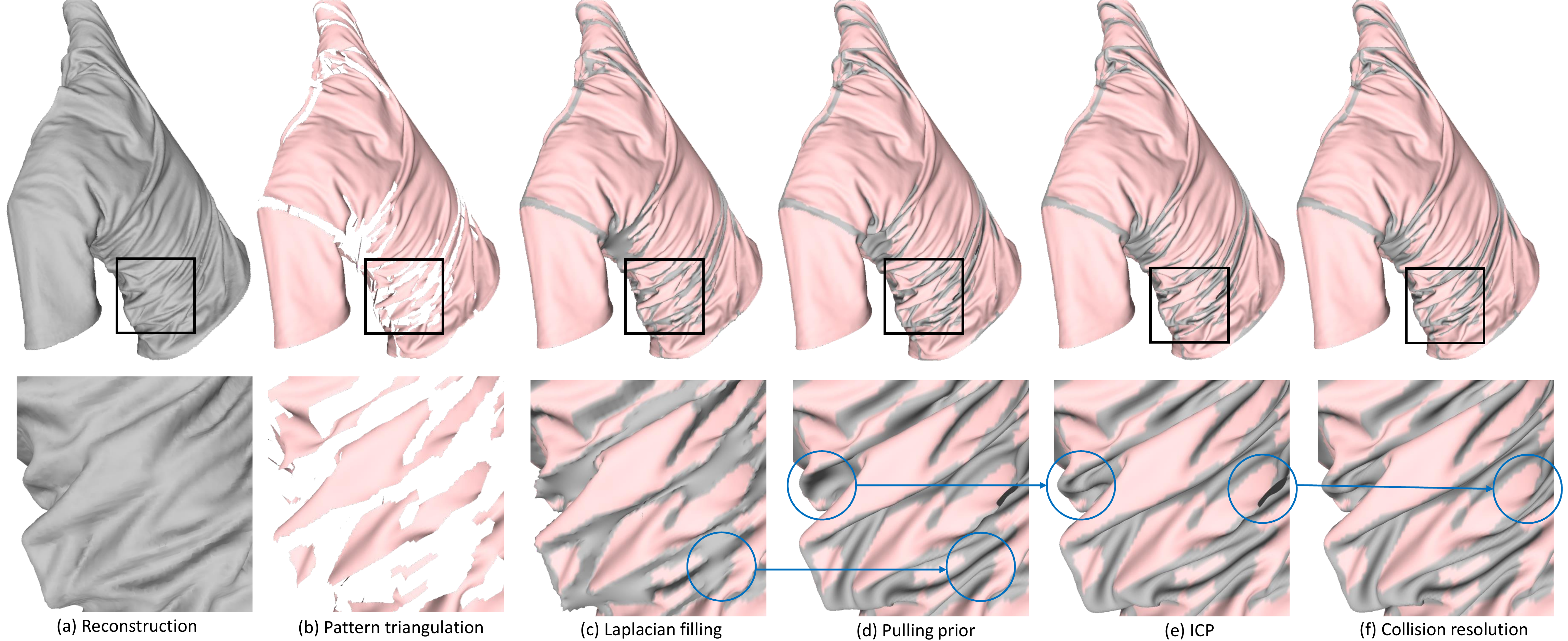}
\caption{\textbf{Multi-view Registration} We implemented an incremental per-frame registration approach to fit multi-view reconstructions  and pattern triangulations. Our deformation state is initialized by from  the triangulated vertices (detected) and Laplacian filling (non detected). We use a pulling prior to deform non detected regions towards occluded-concave sections, and ICP  to cover unregistered reconstruction surface. Finally we run physics based correction module to resolve collisions. We highly regions of improvement after each step of our incremental method.} 
\label{fig:panoptic_registration} 
\end{figure*}

To build realistic priors of garment deformation, we capture precise cloth motion in a multi-view studio and align a template mesh representation of the the garment to the pattern registrations.  

\subsection{Setup}
The capture studio consists of 211 calibrated cameras with varying focal lengths that provide distinct levels of detail of our pattern. 

For our capture, a subject wearing the patterned cloth is located at the center of stage and asked to perform several types of actions. We capture natural cloth deformation produced by deep breathing, realistic wrinkle formation derived by torso motion, and challenging cloth deformation produced by hand-cloth interactions.

\subsection{Pattern Triangulation}
\begin{figure}[h]
\centering 
\includegraphics[width=0.47\textwidth]{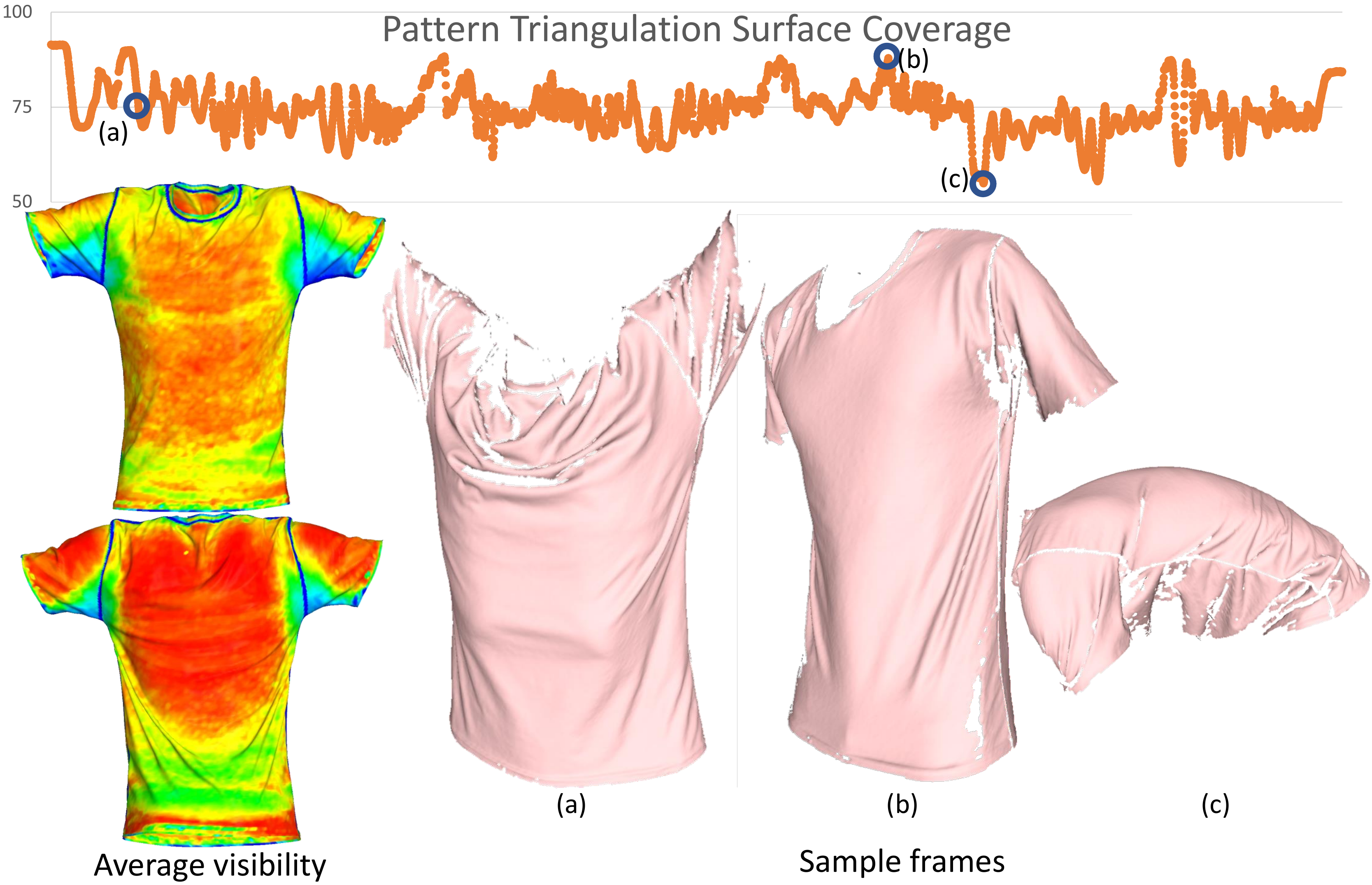}
\vspace{-0.5cm}
\caption{\textbf{Pattern Triangulation.} Our pattern registration and multi-view system, enables a dense registered triangulation. We achieved between $54\%$ and $90\%$ of surface coverage per frame, with average of 75$\%$ (see top plot). We show representative frames attaining mean(a),max(b), and min(c) coverage. We also provide a heatmap illustrating average per vertex visibility.} 
\label{fig:pattern_triangulation} 
\end{figure}

The first step in the panoptic registration is triangulating the multi-view pixel registrations from all the camera views to get the registered point cloud. Each 3D point in the resulting point cloud maps to the pattern square grid coordinates.
This step functions as an independent filtering stage, filtering any remaining outlier pixel registrations from the pattern registration pipeline.
We perform triangulation using the RANSAC algorithm, considering only intersections within $1$ millimeter radius of at least $3$ different rays, as a valid triangulation. The details of the triangulation algorithm are covered in the supplementary. We illustrate the surface coverage provided by the triangulation of the registered pattern in Figure~\ref{fig:pattern_triangulation}.

\subsection{Surface Template Model}

A template mesh $\mathcal{M}$ is constructed to model the state of the full garment surface. Our template mesh enables intuitive mapping between the pattern board, the garment surface, and the UV layout.

The 2D garment pattern on the printed fabric (top row of Figure~\ref{fig:desing_and_fabrication}) provides an initial estimate of the pattern cells that are visible on the shirt. However, some of these cells get occluded after sewing, requiring further refinement based on multiview pattern registration. We construct an initial planar mesh by triangulating these active cells in such a way that cell centers are covered by mesh vertices. This produces a mesh with 5 connected components that have a 1-1 correspondence from vertices to visible pattern cells. To produce a fully connected mesh, we lift our planar components to fit pattern triangulations, and manually extend the mesh tessellation to close seam gaps in regions of good visibility. Our zippered template mesh contains 101286 vertices: 98618 corresponding to visible cell centers and 2668 introduced for seam closure. We define the initial embedding of $\mathcal{M}$ by fitting the tessellation to a T-pose frame and running ARAP\cite{Sorkine2007} refinement to closely preserve isometry to the planar mesh. 

We produce a UV parameterization for the zippered model by extending the initial planar mesh to incorporate the triangles introduced on the seam closure. We use the work by Sawhney~et~al.\,\shortcite{Sawhney2017} to compute a flat extension for the new triangles. The final UV layout  is obtained by arranging the charts to compactly fit in a square. 


 


\subsection{Surface Alignment}

\subsubsection{Deformation Parameterization} 
To align our surface template $\mathcal{M}$ to each frame of the multi-view capture, we use a dense parametric model of surface deformation. The deformation model associates each vertex $v_k\in \mathcal{M}$ with a fixed coordinate frame $F_k$ centered at $v_k$, and a local rigid transformation $\phi_k$. For a given local transformation $\phi_k$, we obtain a global transformation $\tilde{\phi_k} := F_k \circ \phi_k\circ F_k^{-1}$, such that  $\tilde{\phi_k}(v_k)$ corresponds to the deformed vertex position in world coordinates. Given a vector of local transformations $\bm{\phi} := \{\phi_k\}$, we denote the fully deformed mesh as $\mathcal{S}(\bm{\phi})$. In particular, $\mathcal{S}(\bm{0}) = \mathcal{M}$. 
\subsubsection{Incremental Optimization}

We run registration optimization in several stages: first, we optimize each frame independently, from the rest state $\mathcal{M}$ to a deformation state $\mathcal{S}$. Then, we run refinement passes to update $\mathcal{S}$ using the previous pass state (i.e., a Jacobi style update). This enables full parallelism on the capture processing.
We empirically observed that this incremental approach provides a satisfactory trade-off between performance and fitting quality compared with direct optimization from the rest state.

\textbf{Initialization}: The first step of the optimization is to optimize $\bm{\phi}$ such that $\mathcal{S}(\bm{\phi})$ matches well to vertices with valid triangulated positions. We first construct a mesh $\mathcal{P}$ from triangulated vertices in the cloth pattern. We construct $\mathcal{P}$ to have identical topology than $\mathcal{M}$. The detected vertices in $\mathcal{P}$ are assigned to triangulated positions, while the positions of the remaining non-detected vertices are set via Laplacian filling~\cite{Meyer2003}.

 For each vertex $v_k \in \mathcal{M}$ we initialize $\phi_k$ according to the as-rigid-as-possible transformation~\cite{Sorkine2007} mapping the neighborhood of $v_k$ in $\mathcal{M}$ to the respective neighborhood in $\mathcal{P}$. 

Then, we refine $\bm{\phi}$ by jointly minimizing the detection and distortion terms, namely $\Edet$ and $\Edist$.
$\Edet$ is the data term with triangulated vertex positions, and is formulated as
\begin{align}
\Edet (\bm{\phi}) & :=  \| \mathcal{P} - \mathcal{S}(\bm{\phi}) \|^2 . \label{eq:fitting_detection} 
\end{align}

$\Edist$ is the regularization term measuring mesh distortion.
Inspired by Sumner et al.~\shortcite{Sumner2007}, we measure local distortion by comparing the action of transformations defined at adjacent vertices of the mesh. Given edge $e_{kl} \in \mathcal{M}$, we denote by $\{s\}_{kl}$ a set of distinctive points associated to edge $e_{kl}$ (in practice we use the midpoint, and edge corners). We compute the distortion error as,
\begin{align}
\Edist (\bm{\phi}) & := \sum_{e_{kl} \in \mathcal{M}} \| \tilde{\phi_k}(\{s\}_{kl}) - \tilde{\phi_l}(\{s\}_{kl}) \|^2 \label{eq:dense_rigidity} .
\end{align}
This produces an initial deformed mesh $\mathcal{S}_0$ that closely approximates $\mathcal{P}$ and further corrects triangulation outliers due to $\Edist$. 

After this initialization step, we freeze all the valid detected vertices, and only optimize for the non-detected ones in all the subsequent stages. In Figure~\ref{fig:panoptic_registration}, the vertices in the red regions are frozen by the optimization and we only optimize for the deformation parameters of the gray regions. This allows to accelerate the remaining optimization steps due to the sparsity of non-detected vertices.

\textbf{Optimizing Non-Detected Vertices}:
As non-detected vertices are likely occluded, we introduce the following prior: non-detected vertices are more likely to belong to occluded-concave sections (e.g., occluded sections in wrinkles or armpits) than to visible-concave ones. We consider this prior in optimization by formulating the \emph{pulling constraint} on the non-detected vertices, as
\begin{align}
\Epull (\bm{\phi}) := \| (\mathcal{S}_{0} - \epsilon N(\mathcal{S}_{0}) - \mathcal{S}(\bm{\phi}) \|^2.
\end{align}
This pulling constraint regularizes each non-detected vertex to move in the opposite direction of its normal $N$ computed from $S_0$ with a magnitude $\epsilon$. 
%
We also add an additional data term $\Erecon$ with 3D scan mesh $\mathcal{R}$ computed by Yu~et~al.'s method~\shortcite{yu2021function4d}, as some of non-detected vertices could be registered to multi-view stereo.
$\Erecon$ is formulated as 
\begin{align}
\Erecon (\bm{\phi}) := \| \text{ICP}(\mathcal{S};\mathcal{R}) - \mathcal{S}(\bm{\phi}) \|^2, \label{eq:fitting_closest}
\end{align}
where $\text{ICP}(\mathcal{S};\mathcal{R})$ is the (reduced) iterative closest point association from $\mathcal{S}$ to $\mathcal{R}$. More precisely, for each vertex in $\mathcal{R}$ we compute the closest in $\mathcal{S}$, and define a partial correspondence map from $\mathcal{S}$ to $\mathcal{R}$ by applying reduction.

\textbf{Solving Self-Intersections}:
The above optimization steps do not guarantee intersection-free meshes. To alleviate mesh collisions, we employ a physics-based correction module that locally corrects intersecting regions by running a cloth simulation~\cite{stuyck2018cloth} while leaving non-intersecting regions unmodified. We first detect intersecting regions using a surface area heuristic~\cite{baraff2003untangling}, and then run cloth simulation where we apply corrective forces to these regions to resolve self-intersection. 


Figure~\ref{fig:panoptic_registration} illustrates the mesh state at different stages of the optimization.

\textbf{Temporal refinement}: The final step of cloth surface alignment is to make meshes temporally smooth, thus optimizes non-detected vertices via $\Erecon$, $\Edist$ and the smoothness constraint $\Esmooth$ formulated as 
\begin{align}
\Esmooth (\bm{\phi}_t) := \| \mathcal{S} (\bm{\phi}_t) -\frac{1}{4}(\mathcal{S}'_{t-1} + 2\mathcal{S}'_{t} +\mathcal{S}'_{t+1}) \|^2,
\label{eq:fitting_smoothness_damped}
\end{align}
where $\mathcal{S}'$ is a mesh state after the step of solving self-intersection and $t$ is a frame index.
We observed that this smoothness constraint could avoid oscillatory states that we saw with a typical smoothness constraint formulated as $\| \mathcal{S}(\bm{\phi}_t) -\frac{1}{2}(\mathcal{S}'_{t-1} +\mathcal{S}'_{t+1}) \|^2$.
Note that this final optimization can run in parallel because $\Esmooth$ depends on $\mathcal{S}'$ instead of $\mathcal{S}$.

We show the final result of surface alignment on a few frames of the capture in Figure~\ref{fig:selected}. We refer to the supplementary video for further qualitative evaluation of the multi-view aligned meshes.








\begin{figure*}[t]
\centering 
\includegraphics[width=\textwidth]{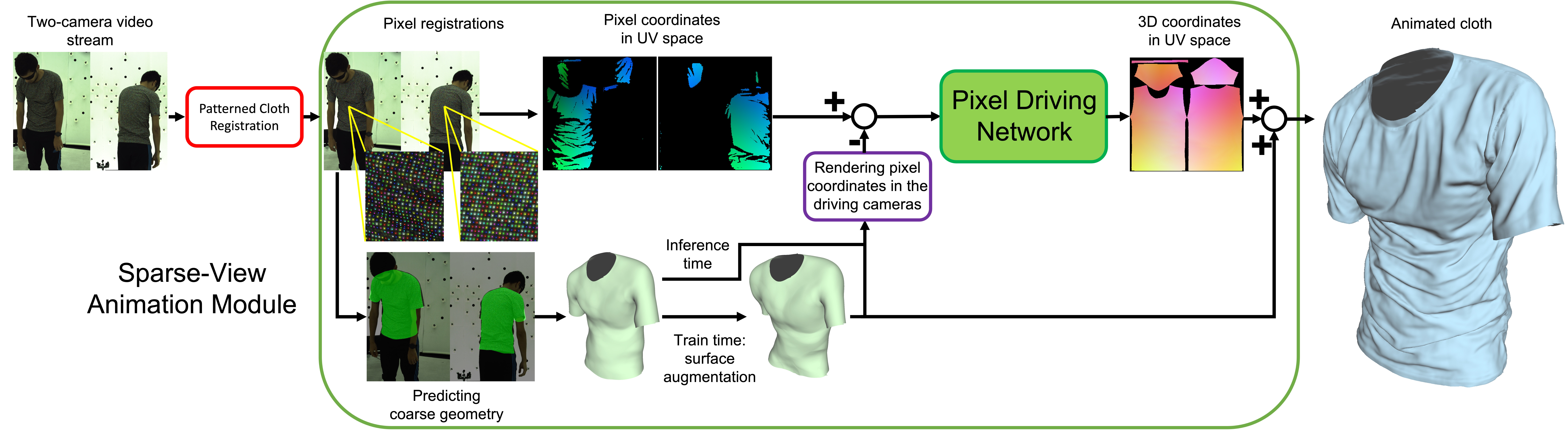}
\caption{\textbf{Garment Avatar Model}. Our model receives a two-camera video stream as an input and generates the animated cloth. The first stage is pattern registration of the input frames, producing pixel registrations for each camera. Then, we use the pixel registrations as input to our \textbf{Sparse-View Animation} module.
First, the pixel registrations are mapped from the image space to the UV space, serving as an input branch to our Pixel Driving network. Additionally, we predict the coarse geometry that aligns with the pixel registrations, generating a reference signal of the pixel coordinates in the UV space of the coarse geometry as rendered in each camera.
To get the input to the network, we subtract the reference signal from the primary signal. To get the final cloth reconstruction, we add the 3D coordinates predicted by the Pixel Driving network as an offset on top of the coarse geometry's 3D coordinates.} 
\label{fig:animation}
\end{figure*}


\section{Sparse View Animation}


\label{sec:animation}
This section introduces our Garment Avatar model, capable 
of producing highly realistic cloth reconstructions from sparse image 
inputs. The model is data-driven, leveraging our accurate cloth deformation data to generate faithful animations even in settings when only two non-overlapping image views are available. We illustrate our model in Figure~\ref{fig:animation}.


\subsection{Input}
We use the video streams of two cameras as input to the \textit{Garment Avatar} model, but our model can trivially be extended to support a different number of views. Note that we assume the cameras are fixed in space, and the calibration is known. 

First, we reuse the pattern registration module to register 
the pixels in the frames of our driving cameras. Each detected point has two attributes 1) a pixel coordinate and 2) a UV coordinate. The pixel registrations for the driving cameras in the image space are the input to the \textbf{Sparse-View Animation} module, which is a significant part of our \textit{Garment Avatar} model.

Then, for each frame $t$, and for each camera $c$, we construct the pixel coordinates signal in the UV domain: $p_{c, t}: \mathcal{U} \rightarrow \mathbb{R}^2$, where $\mathcal{U}$ denotes the UV space.

These pixel coordinate signals in the UV domain serve to generate the input to our pixel-driving network. 

\subsection{Pixel Driving Network}

We consider the two-camera setup with maximum possible coverage, which leads to 
minimal overlap between pattern detections across the views.
However, in practice even the union of the signals' support leaves a significant 
part of the cloth region uncovered.
In other words, each mesh vertex is visible only from a single view at most,
some of the mesh vertices are completely occluded in all the views, and thus
traditional triangulation can not be used directly. 
%
%
We approach this challenge as a special case of an inpainting problem.
Namely, unlike a standard inpainting use-case, the unique feature of our setting is 
that the different camera channels have very limited overlap. 
%
To this end, we stack the pixel coordinates signals of both camera channels
into a tensor containing the joint camera observations 
$P_t \in \mathbb{R}^{U\times V \times 2C}$, 
where $C = 2$ is the number of cameras.
Importantly, the pixel coordinates are normalized with respect to our 
coarse kinematic model, described below.
We then pass this tensor as an input to the pixel driving network - UNet-based architecture, 
which has shown its effectiveness for 
inpainting and image translation problems~\cite{isola2017image,yan2018shift}. 
The pixel driving network predicts 3D coordinates in UV-space $R_t \in  \mathbb{R}^{U\times V \times 3}$.
In practice, the network produces displacements to the coarse geometry of 
the coarse kinematic model to ensure generalization to unseen poses.

\subsection{Coarse Geometry Approximation}

We obtain a coarse approximation for the animated cloth by fitting a kinematic model to the pixel projection seen in each driving camera. Specifically, we render the pixel coordinate signal of the coarse geometry in each driving camera, subtracting it from our pixel registration signal. At the output, we add the predicted 3D coordinates by the network to the 3D coordinates of the coarse geometry. This way, our network maps the offset signal in pixel coordinates to offset signal in world coordinates. Whenever a pixel coordinate is not available at the input channels, we assigned zero value in the corresponding UV coordinate. With this normalization, the input and the output signal are expected to distribute roughly like Gaussian random variables.

\subsubsection{Kinematic Model}
\begin{figure}[h]
\centering 
\includegraphics[width=0.47\textwidth]{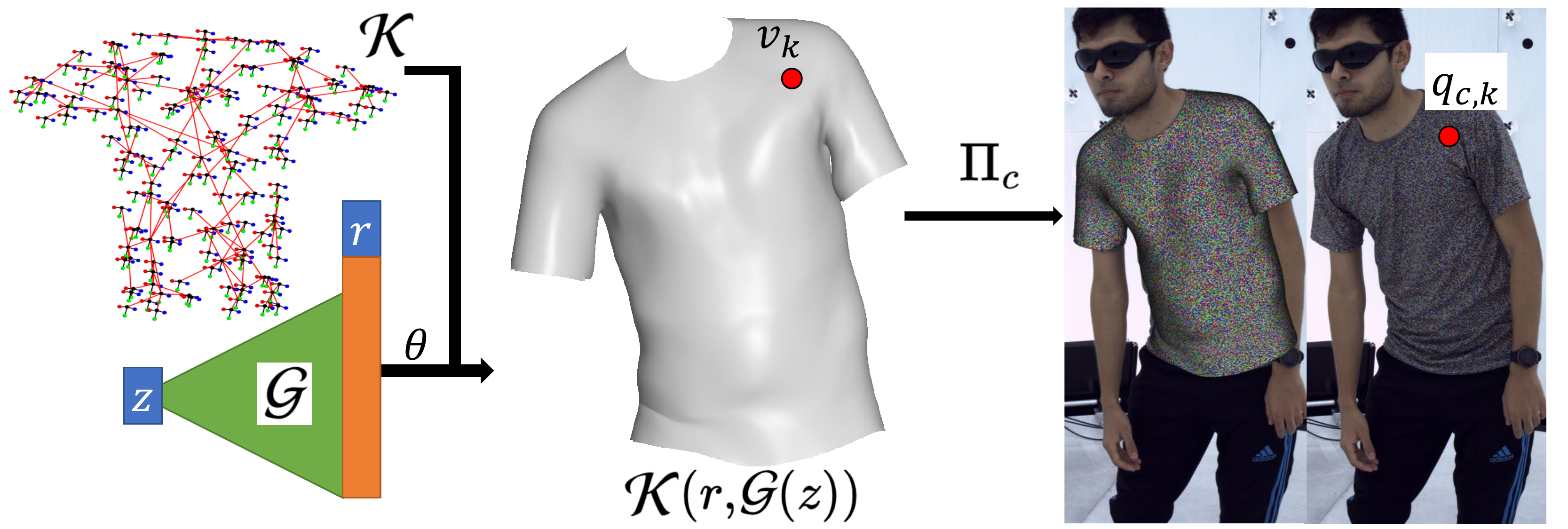}
\caption{\textbf{Coarse Deformation Fitting}. A generative pose model $\mathcal{G}$ is trained from the multi-view aligned meshes to represent the space of coarse cloth deformations on skeleton $\mathcal{K}$. We use this deformation prior to robustly estimate the coarse estate of our cloth from sparse view points.} 
\label{fig:coarse_approximation} 
\end{figure}

A common approach to construct kinematic models for garments is by restricting to certain components of the full body models. For instance, kinematic models for shirts are commonly generated by restricting to torso and upper arms. This provides a good prior for cloth deformation based on body kinematics, but it is only valid for tight cloth deformations. Given the motion complexity we aim to model, and the dense detection provided by our pattern, we created a kinematic with additional flexibility. Our kinematic model $\mathcal{K}$, shown at top left of Figure \ref{fig:coarse_approximation} is a hierarchical deformation graph (i.e., a skeleton) with 156 nodes. We use a procedural approach for its construction that can be applied to other types of garments. Please refer to the supplemental material for details on its construction.

The kinematic model is animated as in standard skeletons\cite{loper2015smpl}: local rigid transformations $\bm{\theta} = \{\theta_i\}$ defined at the nodes of the skeleton, are propagated through the kinematic chain, and vertex positions are computed by linear blend skinning. We denote the skinned mesh by $\mathcal{K}(\bm{\theta})$.  


\subsubsection{Generative Coarse Deformation Model}

To get realistic cloth deformation,
we learn a generative parameter model for $\mathcal{K}$ from the training set. First, we compute deformation parameters $\bm{\theta}_t$ for each multi-view aligned surface $\mathcal{S}_t$ in the training set by minimizing vertex and distortion error:
\begin{equation}
\label{eq:skeleton_fitting}
\bm{\theta}_t := \argmin ||\mathcal{S}_t - \mathcal{K}(\bm{\theta})||^2 + \lambda  E_{\text{Distortion}}(\bm{\theta})
\end{equation}

Here we use the same distortion as in Equation \ref{eq:dense_rigidity}, defining adjacency of skeleton nodes by overlapping skin support.


The set of deformation parameters $\Theta = \{\bm{\theta}_t\}$ characterize the space of realistic cloth deformations given by our training set. We normalize these deformation parameters by removing the root transformation. 
A variational autoencoder-based generative network is trained on this normalized parameters,
similar to \cite{SMPL-X:2019}. Although we supervise the reconstruction error in the deformed mesh space rather than in the parameter space.

We denote the generative parameter model by $\mathcal{G}$. It is designed as an MLP with two hidden layers mapping a 32-dimensional latent code to the 1085 parameter space (rigid transformation of all nodes but the root) of the kinematic model.

\subsubsection{Coarse Deformation Fitting}

 Given a vertex $v_k \in \mathcal{M}$ and camera $c \in C$, we denote by $w_{c,k}$ to the binary value specifying the vertex visibility, and $q_{c,k}$ the pixel coordinate of its corresponding detection (if any). We denote $\Pi_c$ the camera projective operator. Please refer to Figure \ref{fig:coarse_approximation} for further illustration on the notation.

We compute the coarse cloth deformation by optimizing for root transform $r^{*}$ and latent code $z^{*}$, to fit image detections,
\begin{equation}
\label{eq:projection_fitting}
r^{*},z^{*} = \argmin \sum_{c\in C}\sum_{k\in \mathcal{M}} w_{c,k}||\Pi_c(\mathcal{K}(r,\mathcal{G}(z))(v_k)) - q_{c,k}||^2
\end{equation}

We use $\mathcal{K}(r^{*},\mathcal{G}(z^{*}))$ for signal normalization in the Pixel Driving Network. 

\subsection{Surface Augmentation}
To make our driving network insensitive to the precise shape of the coarse geometry, we augment the coarse surface produced by the kinematic model at training time. We deform the coarse geometry using a deformation space spanned by the Laplace-Beltrami eigenfunctions, using a randomized filter to select their linear combination.
The deformation is defined as the displacement along the surface normal modulated by the random scalar function.
We expect the coarse geometry to be projected close to the pixel detections.
Therefore, the most significant ambiguity is likely to be along the local vector parallel to the surface normal. Since the kinematic model produces surfaces with coordinate functions of limited frequency, we used an exponential function as the spectral amplitude to decay the high-frequency modes of the displacement function. We used Bernoulli random variables for the filter coefficients to mix different modalities for each augmentation instance. 
Finally, we scale the resulting displacement function with a Gaussian random variable to allow displacements of varying magnitudes. More precise, our displacement function is:
\begin{equation}
    D(x) = A \sum_{n=1}^{\infty} q_ne^{-\alpha n} \phi_n(x), x \in \mathcal{S}
\end{equation}
where, $\phi_n$ are the corresponding Laplace-Beltrami functions on the coarse surface $\mathcal{S}$, $\alpha=50$ is the filter decay rate in the spectral domain, $A \sim \mathcal{N}(0, \sigma=20mm)$ is the random scale, and $q_n \sim Bernoulli(p=1/2)$ are the random coefficients selecting the different modes.
To a good approximation, $\mathcal{M}$ is isometric to the coarse surface $\mathcal{S}$ and Laplace-Beltrami eigenfunctions are isometry invariant. Therefore, we calculated Laplace-Beltrami eigenfunctions once on the template mesh $\mathcal{M}$, and map them to the coarse geometry.
At inference time, the coarse geometry is passed directly without deforming it. 

\subsection{Training and Inference Setup}
We train our Pixel Driving network, with L2 loss with respect to the ground-truth registered meshes. We apply the L2 loss between the 3D vertex positions and include an L2 between the surface normal vectors to promote the preservation of fine geometric detail. Our training data consists of the first $3500$ frames of the sequence, while the remaining $500$ frames are used as the test set. New deformations never observed in the train set are introduced in the test sequence.

\begin{figure*}
\centering 
\includegraphics[width=\textwidth]{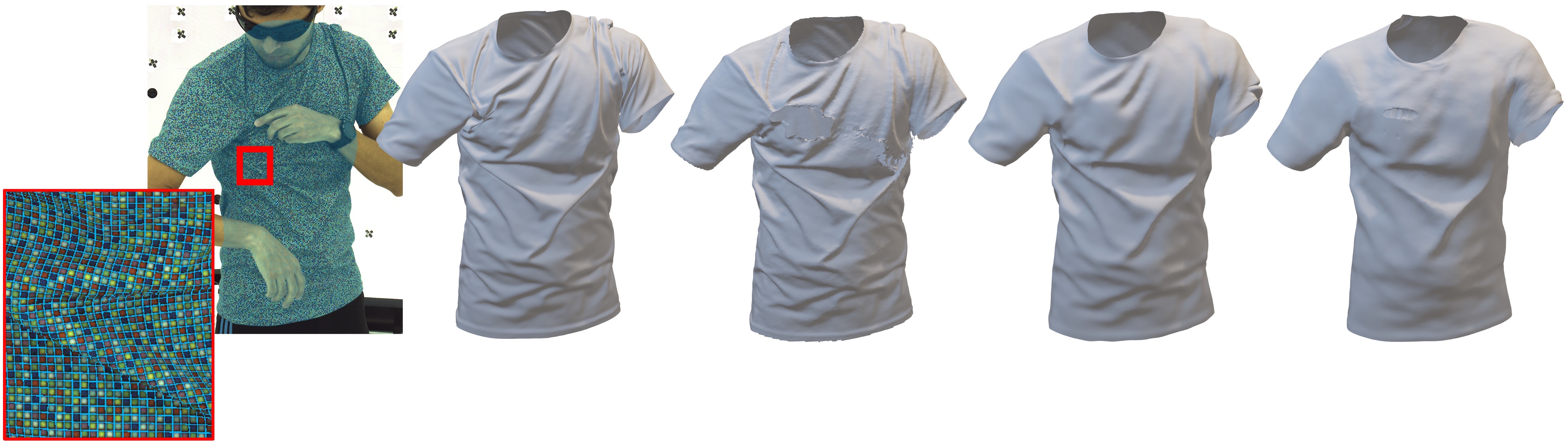}
\caption{\textbf{Registration Qualitative Evaluation.} 1) projecting our registered mesh with a wire-frame texture (aligned with the pattern square grid) to the image 2) Registered mesh geometry: a. ours b. multi-view stereo reconstruction \cite{yu2021function4d} c. body tracking + Non-rigid ICP using \cite{Pons-Moll:Siggraph2017} d. body tracking + Non-rigid ICP using + photometric optimization \cite{xiang2021modeling}.} 
\label{fig:wireframe}
\end{figure*}


\section{Results}
In this section, we showcase our results. We start by evaluating our aligned registered meshes, introducing the comparison methods and the metrics. Then, we display our driving results from sparse observations in two different settings. First, the body pose serves as the driving signal. Then, we use our pattern registrations obtained from two opposite-facing cameras as the driving signal. We show qualitative results and report the numerical driving errors in each case.

\begin{figure*}
\centering 
\includegraphics[width=\textwidth]{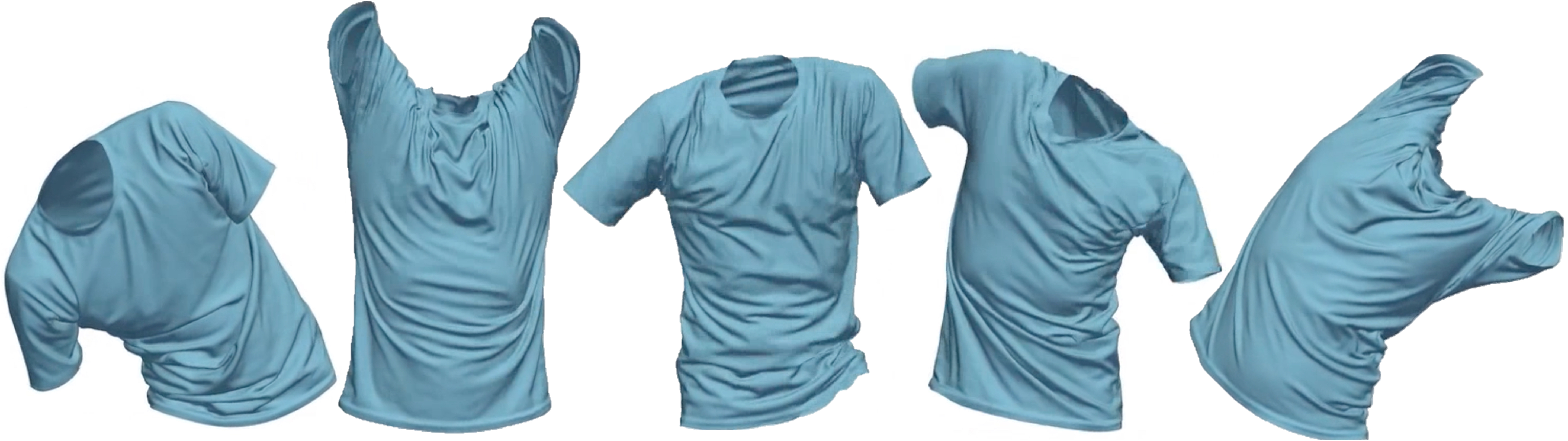}
\caption{Several sample frames from our multi-view aligned mesh sequence.} 
\label{fig:selected}
\end{figure*}

\begin{figure}
\centering 
\includegraphics[width=0.5\textwidth]{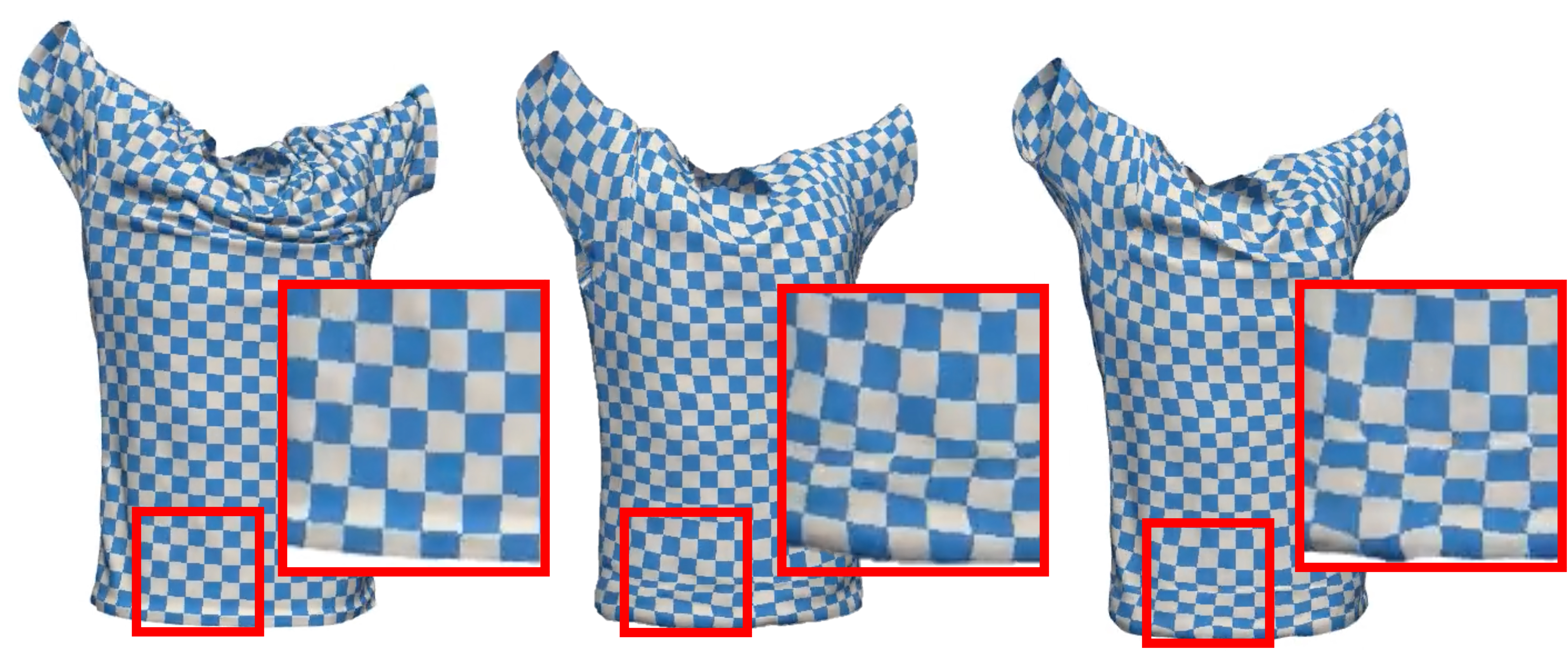}
\caption{Along the deformation sequence, custom texture is overlayed coherently with our registration (1), while with  (2) \cite{xiang2021modeling} and (3) \cite{Pons-Moll:Siggraph2017} the texture jitters and distort} 
\label{fig:texture}
\end{figure}

\subsection{Evaluating the 3D Registration Accuracy}

\begin{figure}
\centering 
\includegraphics[width=0.47\textwidth]{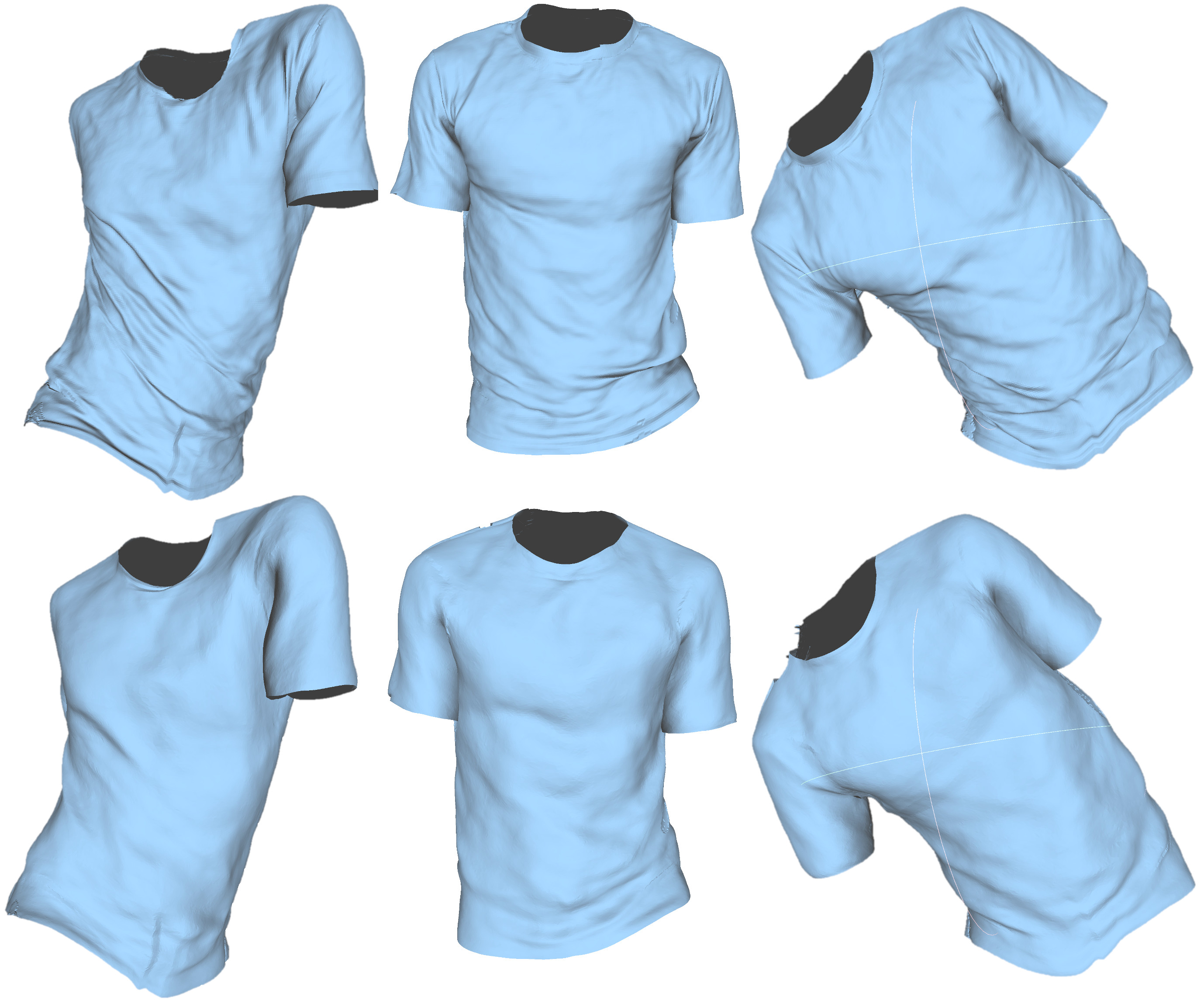}
\caption{Qualitative evaluation of the animated geometry training for pose driving with our registrations (top) and with ClothCap \cite{Pons-Moll:Siggraph2017} registrations (bottom)} 
\label{fig:pose-driving-geometry} 
\end{figure}

\subsubsection{Triangulation}
Figure~\ref{fig:pattern_triangulation} evaluates the coverage of our triangulated point clouds and the average visibility of different cloth regions. 
The precision of the registered point cloud we obtained from RANSAC triangulation is at least $3$ inlier intersecting rays within an intersection radius of $1$ mm. 

\subsubsection{Comparison to Related Work}
Figure~\ref{fig:wireframe} showcases our registration quality by projecting our registered surface to the original image as a wireframe and our reconstructed cloth surface on the left. On the right we show comparisons with related work.
We compare our registered meshes with our implementation of cloth registration pipeline in ClothCap \cite{Pons-Moll:Siggraph2017}\footnote{We use our implementation of ClothCap \cite{Pons-Moll:Siggraph2017} for all the comparison.}. In this method, a kinematic body model is used to track a single-layer surface describing the clothed subject. Using the T-pose frame, the cloth topology is segmented from the single-layer topology and used to define a cloth template.
Then, this cloth template is registered to the clothing region of the single-layer surface using non-rigid Iterative Closest Point (ICP). For better initialization, we used Biharmonic Deformation Fields \cite{jacobson2010mixed} to align the boundary between the deformed template and the target mesh such that the interior distortion is minimal as suggested by Xiang et al. \shortcite{xiang2021modeling}.
We also compare our registered meshes to those produced by the inverse rendering optimization stage presented by Xiang et al. \shortcite{xiang2021modeling}, which aimed to refine the non-rigid ICP registration described above. Finally, we compare our reconstructed geometry to the unregistered multi-view stereo reconstruction \cite{yu2021function4d}.

Visual examination of the registration results reveals a few observations. First, the reconstruction quality, especially in the wrinkles, is the best with our registration pipeline, even when compared to the unregistered multi-view stereo reconstruction \cite{yu2021function4d}. Secondly, since our method does not rely on a kinematic body model for its operation, it is stable even when the cloth moves tangentially over the body or detaches from the body.
On the contrary, ClothCap \cite{Pons-Moll:Siggraph2017} relies on the kinematic body model 
and shows artifacts at the sleeve region where the cloth is not perfectly overlapping with the body model.
More severe instability due to tangential movement appears in the supplementary video. Similarly, Xiang et al. \shortcite{xiang2021modeling} also rely on the body kinematic model for initialization, but later it refines the registration using inverse rendering optimization. This stage helps fix the initial artifacts but is vulnerable in the invisible cloth regions, such as the armpits and the areas occluded by the body parts, producing artifacts visible in Figure~\ref{fig:wireframe}~and~\ref{fig:texture} and more severe artifacts visible in the supplementary video. On the other hand, our registration pipeline has smooth results, even in the invisible regions. 

\subsubsection{Error Metrics}

Common error metrics such as pixel reprojection and rendering errors are hard to transfer from one capture system to another. To provide reproducable metrics that are invariant to the specific multi-view camera setting, we propose two novel absolute measurements.
The first is the correspondence drift rate in [mm/sec] units. The second is the average geodesic distortion in [mm] over the sequence of frames.
Table~\ref{table:registration errors} reports these metrics for our work that future work can reference when accurate ground truth is unavailable.

To measure the correspondence drift rate, we unwrap the image texture to the UV domain and measure the optical flow. We translate the optical flow magnitude to correspondence drift rate in [mm/sec] units, using the square size calibration and the known camera frame rate.
We also report the per-frame average geodesic distortion (a.k.a Gromov Hausdorf distortion) w.r.t a reference frame in a T-pose. To this end, we randomly sampled the vertex pairs, measured their pairwise geodesic distance, and compared it to the geodesic distance measured in the reference frame. We used \textit{pygeodesic} implementation of the exact geodesic algorithm for triangular meshes \cite{mitchell1987discrete}.
We don't report the Chamfer distance to the reconstruction surface extracted from the multi-view stream using the work of Yu et al. \shortcite{yu2021function4d} because our reconstruction quality is better than this ground-truth candidate, leaving this measurement too coarse for evaluation.
The table clearly shows that our correspondence drift rate is negligible relative to the compared methods.
Our geodesic distortion is significantly lower. The former is directly related to the drift a texture applied to the geometry will exhibit. The latter is related to the deviation of the reconstructed geometry from the expected one, assuming the cloth surface deforms roughly isometrically.

We measured the registration metric also on a bigger pattern we used to ablate our registration pipeline. The big pattern square size is $4$ mm, compared to the pattern we use of size $2.67$ mm. We hoped those metrics would help compare those patterns, but the errors reported in the table for both patterns deviate from others within the precision of the optical flow and geodesic distortion measurement techniques, which are both approximately $1$ mm. Hence, we conclude that we should devise a measurement with sub-millimetric precision to analyze the difference between them quantitatively, as the standard measures are too coarse.

Finally, Figure~\ref{fig:selected} showcases a few sampled frames from our registered mesh sequence, and Figure~\ref{fig:texture} shows our coherent texture mapping, as opposed to other registration methods for which the texture jitters and distort. We include the full videos in the supplementary material.

\begin{table*}
    \centering
    \begin{tabular}{ |p{8.5cm}||p{5cm}|p{3.5cm}|  }
     \hline
     \multicolumn{3}{|c|}{Registration errors} \\
     \hline
     Method & Correspondence drift rate [mm/sec] & Geodesic distortion [mm] \\
     \hline
     Pattern registration  &  1.50    & 9.27  \\
     Pattern registration  (big pattern) &  0.77    & 9.98    \\
     Body tracking + non-rigid ICP  \cite{Pons-Moll:Siggraph2017} &  72.3     & 25.94   \\
     Body tracking + non-rigid ICP + IR optimization  \cite{xiang2021modeling} & 64.6    & 35.49  \\
     \hline
    \end{tabular}
    \caption{Quantitative evaluation of the registration method}
    \label{table:registration errors}
\end{table*}

\begin{table*}
    \centering
    \begin{tabular}{ |p{9.5cm}||p{3.5cm}|p{3.5cm}| }
     \hline
     \multicolumn{3}{|c|}{Driving errors w.r.t. ground truth registered surfaces} \\
     \hline
     Method & Euclidean distance [mm]& Chamfer distance [mm] \\
     \hline
     Our pixel registration driving (surface kinematic model)  &  2.08    & 1.59   \\
     Our pixel registration driving (LBS model) &  7.22    & 4.05   \\
     Pose driving \cite{xiang2021modeling} trained with our registrations & 15.89    & 5.05    \\
     Pose driving \cite{xiang2021modeling} trained with non-rigid ICP registration     & 22.64    & 5.42    \\
     \hline
    \end{tabular}
    \caption{Driving metric evaluation}
    \label{table:driving errors}
\end{table*}

\subsection{Driving from sparse observations}
\subsubsection{Driving from pose}
\begin{figure}
\centering 
\includegraphics[width=0.5\textwidth]{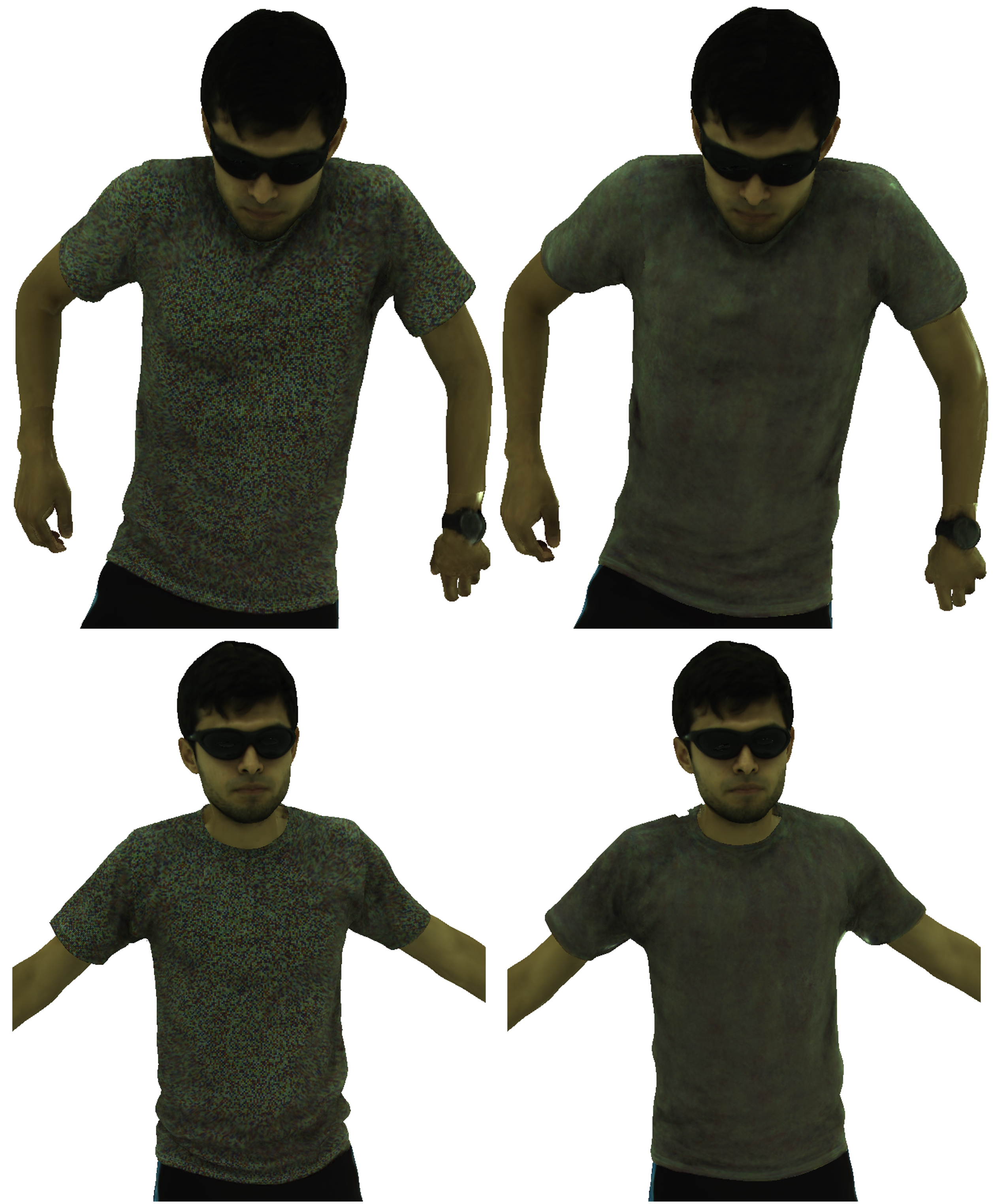}
\caption{Our high-quality cloth registrations improve the performance of the appearance model in the pose driving network. Training for pose driving with our registrations (left), and with the registration pipeline in \cite{Pons-Moll:Siggraph2017} (right)} 
\label{fig:appearance} 
\end{figure}

To prove the claim that the current bottleneck for virtual reality cloth modeling is the lack of accurate cloth registrations, we evaluate the driving method proposed by Xiang et al. \shortcite{xiang2021modeling} once with the non-rigid ICP training data, and once with our cloth registrations. Figure~\ref{fig:pose-driving-geometry} shows the improved wrinkle modeling in the animated geometry and Table~\ref{table:driving errors} reports the corresponding errors in the predicted geometry, showing a clear advantage for the network trained with our accurate registrations compared to the registrations obtained with ClothCap \cite{Pons-Moll:Siggraph2017}.
Alongside the improvement in the geometry, our registrations also promote a significant improvement in the appearance modeling. It resolves the challenging problem that all the codec-avatar methods face when trying to recover both unknown textures and unknown geometry from a given capture. Without accurately anchored geometry, the appearance, optimized by inverse rendering, achieves poor results for non-trivial textures, see Figure~\ref{fig:appearance}. 

\begin{figure*}
\centering 
\includegraphics[width=\textwidth]{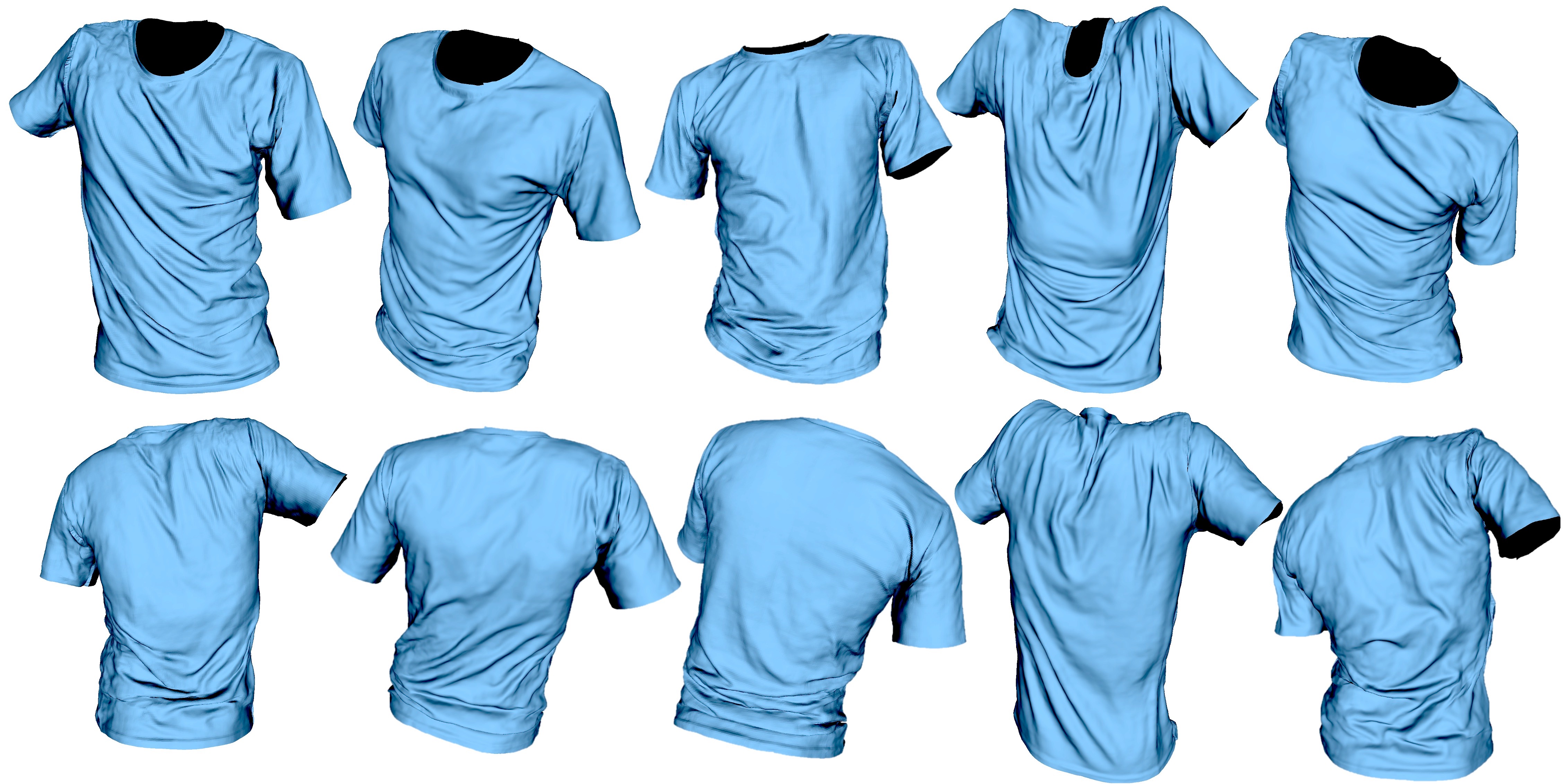}
\caption{A selected set of frames of our cloth sequence animated from two cameras pixel registrations: frontal view (top), and rear view (bottom).} 
\label{fig:pixel_driving} 
\end{figure*}

\begin{figure}
\centering 
\includegraphics[width=0.5 \textwidth]{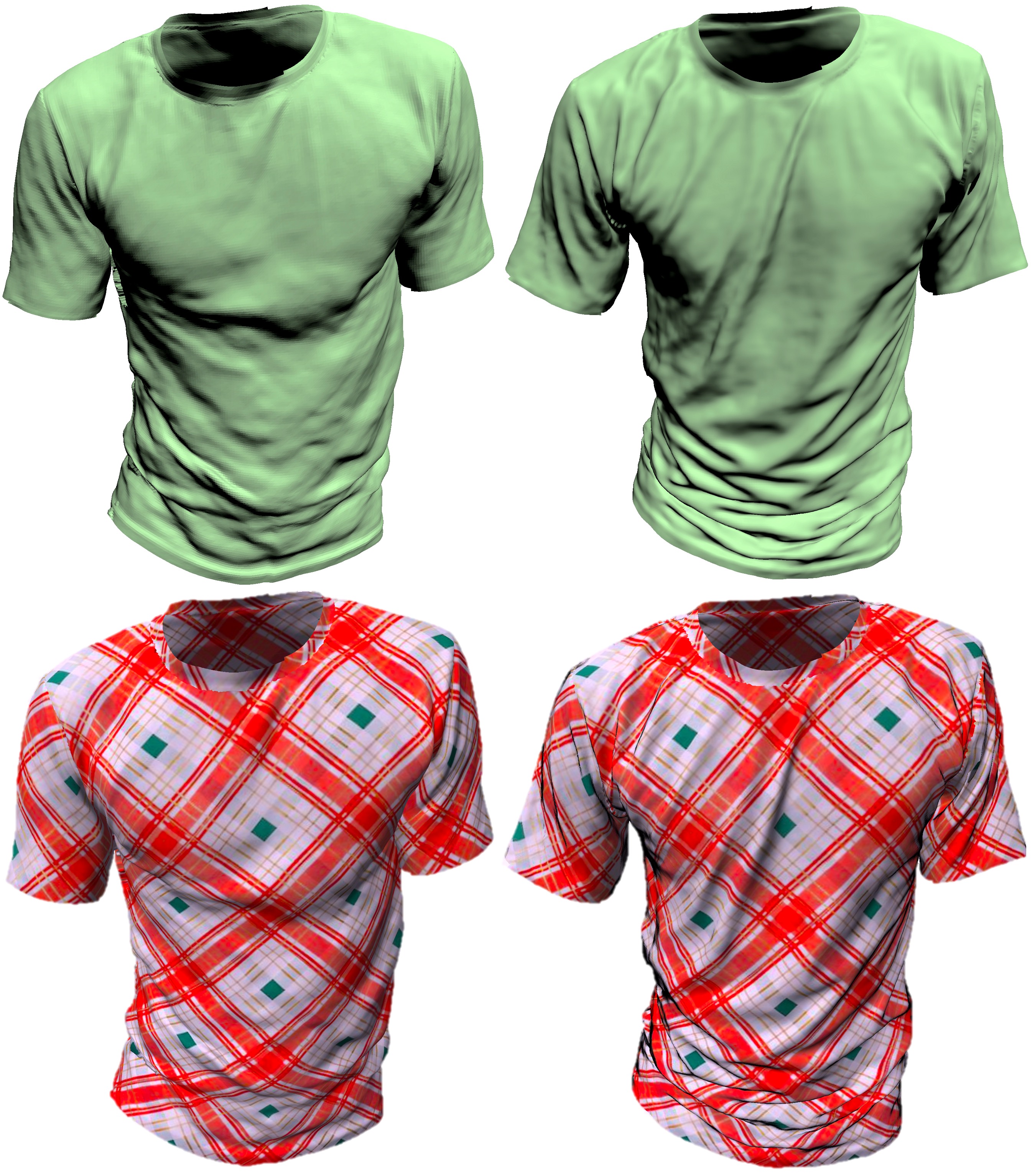}
\caption{Pixel registration driving (right) enhances the driving results' geometric modeling and realistic appearance compared to pose driving (left).} 
\label{fig:qualitative_driving} 
\end{figure}

\subsubsection{Driving from Pattern Registrations}
Here, we showcase the results of our sparse-view animation module trained with our high-quality registered clothes data.
We evaluate the generalization of our driving network for the different deformation sequence that appears in the test set. Figure~\ref{fig:pixel_driving} shows selected sample frames of the resulting animated cloth sequence. Since our method processes a dense driving signal, while pose-driven neural models rely on the sparse skeletal joint parameters, we cannot directly compare them. Whether to apply pose driving or drive from pixel registration depends on the specific application. Figure~\ref{fig:qualitative_driving} shows the level of detail and realism that the pixel registration driving adds to the final driving results. We show the pose-driven network \cite{xiang2021modeling} and our pixel registration-driven model, both trained with our high-quality registration data, differing in the driving mechanism. Table~\ref{table:driving errors} reports the driving errors for each driving method. Finally, to empirically justify our choice of the surface kinematic model, we ablate the kinematic model used to estimate the coarse geometry in our method. We report the driving errors when LBS is used as the kinematic model and when we use our surface kinematic model.

\section{Discussion, Limitations and Future Work}
We proposed a drivable Garment Avatar model generating highly realistic cloth animation from sparse-view captures of patterned cloth. We developed a pattern registration and surface alignment pipeline producing a dynamic sequence of registered meshes representing the cloth deformation in an unprecedented resolution, localization error, and correspondence accuracy significantly better than any prior art. Finally, we demonstrated how are cloth registration technique can be used to improve existing pose-driving methods. Our pattern registration method supports future data generation of various garments and facilitates better geometry and appearance modeling for telepresence. 

Moreover, our approach offers a novel way to capture data to bridge the gap between synthetic simulation and real-world clothes. Applications include enhancing cloth physical simulations \cite{runia2020cloth, jin2020pixel}, facilitating the development of neural models for cloth dynamics \cite{lahner2018deepwrinkles, liang2019differentiable, holden2019subspace} and non-rigid shape correspondence \cite{attaiki2021dpfm, eisenberger2020smooth, bracha2020shape, halimi2019unsupervised, eisenberger2020deep, litany2017deep}, designing precise interfaces for cloth manipulation by robots \cite{bersch2011bimanual, miller2012geometric, strazzeri2021topological}, ground-truth data to generative models for shape completion \cite{bednarik2020shape, halimi2020towards, chi2021garmentnets}, and interpolation \cite{cosmo2020limp, eisenberger2019divergence}.

\vfill
In the future, we plan to pursue the following research directions. The immediate future goal is to optimize our registration pipeline and sparse-view animation module for real-time performance, to support the requirement of a telepresence framework to perform real-time driving. Additionally, we aim to generalize our driving model to support a possibly different camera configuration in inference time. We believe the key is to train a modified version of our Pixel Driving network that utilizes the 3D direction of the ray through each pixel rather than only the pixel coordinates. To resolve the geometrical artifacts along the seams, we plan to implement our pixel-driving network in the mesh domain, utilizing mesh convolutions and mesh pooling and un-pooling layers \cite{zhou2020fully, hanocka2019meshcnn} that will replace the current layers in the UV domain. Our driving model operates frame-wise, exhibiting slight jitter in the resulting sequence. We plan to overcome this issue by devising a recurrent architecture for the cloth animation \cite{santesteban2019learning, santesteban2021self}.
We plan to register the corners with the centers in future captures, consequently doubling the mesh resolution.
We plan to devise an automatic template generation procedure for the multi-view surface alignment to make our data generation scalable. To this end, we would like to utilize geometrically invariant signatures \cite{grim2016automatic, halimi2018self} to recover the global topology of the sewing pattern. Then we could use invariant parameterizations \cite{halimi2019computable} to calculate the pointwise correspondence along the sewing lines. Finally, we explored the geometric direction, learning to simulate shape deformations from sparse observations, leaving the appearance modeling completely undiscovered. However, we evidenced a highly coherent texture signal when projecting the camera images to our accurate geometry. The latter implies that practically, we can learn the function that maps geometry to shading from pure observations.

\section*{Acknowledgements}
We thank Printful Inc. and Luz Diaz-Arbelaez for cloth fabrication. Alec Hnat, Autumn Trimble, Dani Belko, Julia Buffalini, Kevyn McPhail, and Rohan Bali for assistance on data capture. Aaqib Habib and Lucas Evans for supervising annotations. Kaiwen Guo and Yuan Dong for contributions to data pre-processing. Oshri Halimi is funded by the Israel Ministry of Science and Technology grant number  3-14719.

\bibliographystyle{ACM-Reference-Format}
\bibliography{bib}


\end{document}